\pdfoutput=1

\documentclass[11pt]{article}

\usepackage{ACL2023}

\usepackage{times}
\usepackage{latexsym}

\usepackage[T1]{fontenc}

\usepackage[utf8]{inputenc}

\usepackage{microtype}

\usepackage{inconsolata}

\usepackage{color}
\usepackage{colortbl}
\usepackage{graphicx}
\usepackage{multirow}
\usepackage{makecell}
\usepackage{float}
\usepackage{subfig}
\usepackage{bm}
\usepackage{upgreek}
\usepackage{booktabs}
\usepackage{amsfonts,amssymb}
\usepackage{amsmath}
\usepackage{bbding}
\usepackage{array}

\usepackage{fontawesome} 
\usepackage[most]{tcolorbox}
\usepackage{url}

\makeatletter
\newcommand{\thickhline}{%
    \noalign {\ifnum 0=`}\fi \hrule height 1pt
    \futurelet \reserved@a \@xhline
}
\newcolumntype{"}{@{\hskip\tabcolsep\vrule width 1pt\hskip\tabcolsep}}
\makeatother
\usepackage[linesnumbered,ruled,vlined]{algorithm2e}
\SetKwComment{Comment}{$\rhd$}{}
\AtBeginDocument{%
  \providecommand\BibTeX{{%
    \normalfont B\kern-0.5em{\scshape i\kern-0.25em b}\kern-0.8em\TeX}}}

\newcommand{\ie}{\textit{i.e.}}
\newcommand{\eg}{\textit{e.g.}}

\newcommand{\blue}{\textcolor{blue}}

\newcommand{\smallsnow}{\includegraphics[height=0.7em]{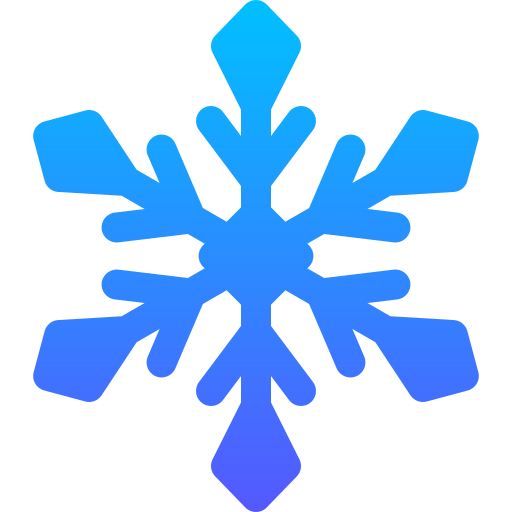}}
\newcommand{\smallfire}{\includegraphics[height=0.7em]{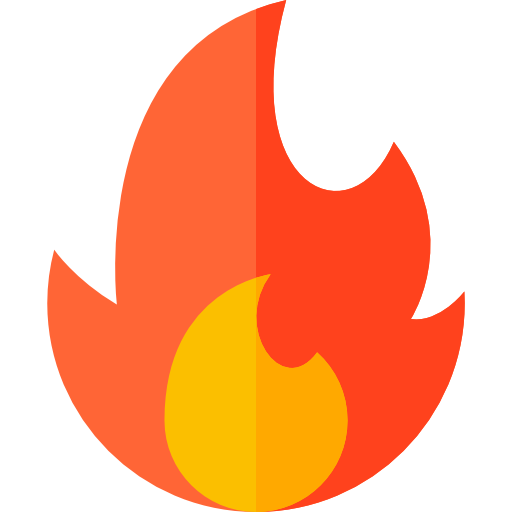}}

\definecolor{veronica-red}{RGB}{196,30,58}

\title{Self-supervised Quantized Representation for Seamlessly Integrating Knowledge Graphs with Large Language Models}

\author{Qika Lin$^{\heartsuit}$, Tianzhe Zhao$^{\diamondsuit}$, Kai He$^{\heartsuit}$, Zhen Peng$^{\diamondsuit}$, Fangzhi Xu$^{\diamondsuit}$ \\
{
\bf Ling Huang$^{\heartsuit}$, Jingying Ma$^{\heartsuit}$, Mengling Feng$^{\heartsuit}$}\\
$^\heartsuit$National University of Singapore\;\;
$^\diamondsuit$Xi'an Jiaotong University\\
  \texttt{qikalin@foxmail.com}\quad \texttt{ephfm@nus.edu.sg}
  }

\begin{document}
\maketitle
\begin{abstract}
Due to the presence of the natural gap between Knowledge Graph (KG) structures and the natural language,
the effective integration of holistic structural information of KGs with Large Language Models (LLMs) has emerged as a significant question.
To this end, we propose a two-stage framework to learn and apply quantized codes for each entity, aiming for the seamless integration of KGs with LLMs.
Firstly, a self-supervised quantized representation (SSQR) method is proposed to compress both KG structural and semantic knowledge into discrete codes (\ie, tokens) that align the format of language sentences.
We further design KG instruction-following data by viewing these learned codes as features to directly input to LLMs, thereby achieving seamless integration.
The experiment results demonstrate that SSQR outperforms existing unsupervised quantized methods, producing more distinguishable codes.
Further, the fine-tuned LLaMA2 and LLaMA3.1 also have superior performance on KG link prediction and triple classification tasks, utilizing only 16 tokens per entity instead of thousands in conventional prompting methods.
\end{abstract}

\section{Introduction}
\label{sec_intro}

Large Language Models (LLMs), such as LLaMA~\cite{touvron2023llama,touvron2023llama2} and GPT-4~\cite{achiam2023gpt}, are initiating considerable transformations within the fields of artificial intelligence (AI) and natural language processing (NLP).
They have achieved substantial success~\cite{peng2023instruction,DBLP:conf/acl/WangS0G24,DBLP:conf/acl/XuWSRYYLQ024}, and thus, have been regarded as potential pathways towards achieving the ultimate goal of artificial general intelligence~\cite{yang2024harnessing}.
However, the specific training strategies employed by LLMs render them black-box models and struggle to retrieve the relevant facts necessary for the correct answer~\cite{pan2024unifying}, resulting in low performance in complex reasoning scenarios~\cite{xu2023large,DBLP:conf/acl/XuLZ0024}.
Furthermore,
knowledge hallucination becomes a serious issue, which may generate wrong statements that conflict with reality~\cite{bang2023multitask,ji2023survey}.
It presents considerable risks, particularly in specialized fields like law~\cite{cui2023chatlaw} and healthcare~\cite{LIN2025102795,he2025survey}.





\begin{figure}[t]
\centering
\includegraphics[width=1\linewidth]{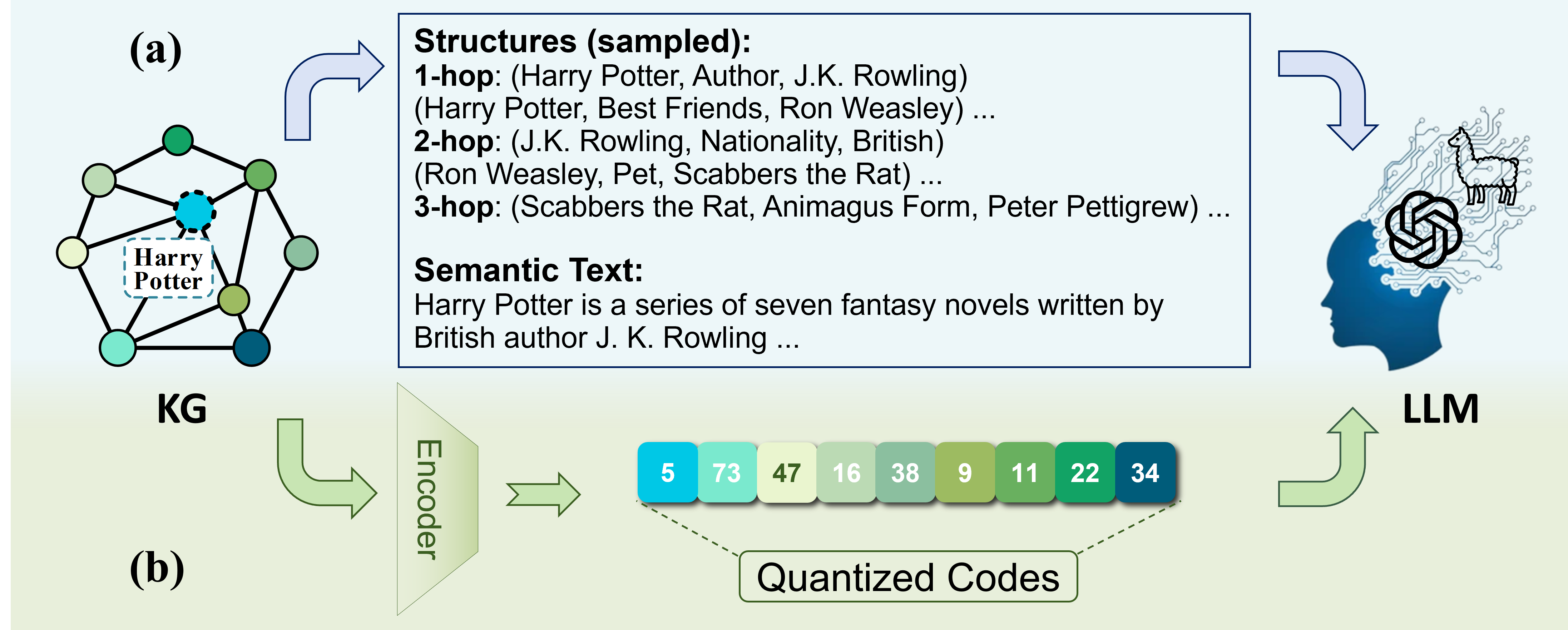}
\setlength{\abovecaptionskip}{-0.2cm}
\setlength{\belowcaptionskip}{-0.5cm}
\caption{Illustration of different strategies to integrate KGs with LLMs. (a) The direct method utilizes (sampled) graph structures and semantic text as inputs.  (b) Our method for seamlessly integrating KGs with LLMs using learned quantized and discrete codes.}
\label{fig_intro}
\end{figure}

Knowledge Graphs (KGs), also known as knowledge bases,
organizes massive amounts of factual knowledge in a structured and interpretable manner by the triple form of (\emph{subject}, \emph{relation}, \emph{object}).
They can serve as a vital supplement to LLMs~\cite{pan2024unifying}, providing an alternative way to address hallucinations and generate more precise answers using continual fine-tuning~\cite{DBLP:conf/mm/00090GX0C24,hron2024training} or retrieve-based reasoning~\cite{DBLP:conf/iclr/SunXTW0GNSG24,tan2024paths,DBLP:conf/acl/ZhangSGXLL24}.
However, the KGs' structure is in a graph form, which markedly differs from the discrete token format of the natural language in LLMs.
Thus, due to the presence of this natural representation gap,
the effective integration of comprehensive structural information of KGs with LLMs has emerged as a significant question.

\begin{figure}[t]
\centering
\includegraphics[width=1.0\linewidth]{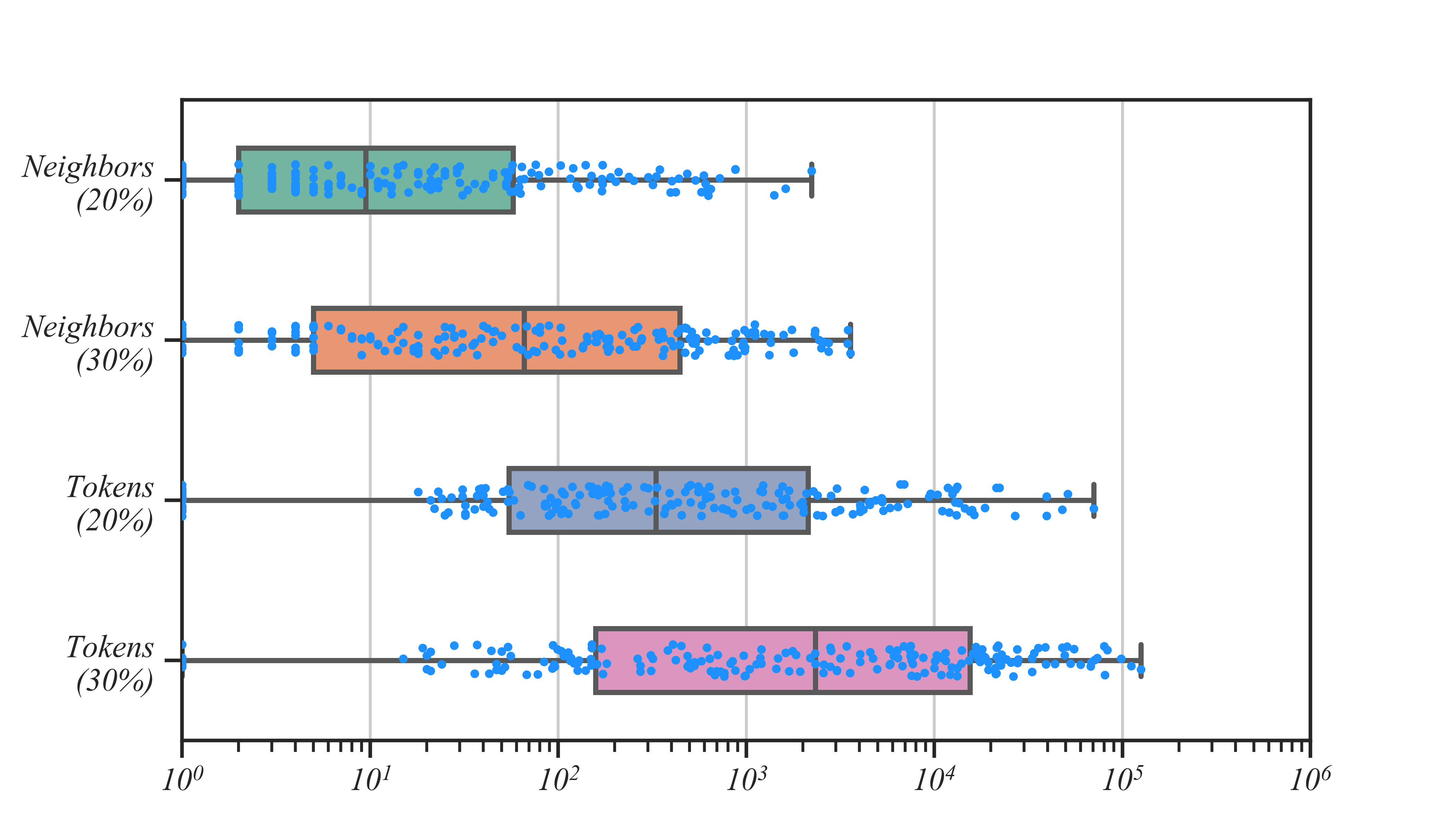}
\setlength{\abovecaptionskip}{-0.5cm}
\setlength{\belowcaptionskip}{-0.5cm}
\caption{The statistics of 2-hop sampled neighbors and needed tokens (by LLaMA2) for entities in FB15k-237.
}
\label{fig_intro_fb}
\end{figure}

As shown in Figure~\ref{fig_intro} (a), one straightforward method involves converting relevant triples into textual prompts and then feeding them into LLMs, combined with semantic text.
This simple strategy would necessitate a substantial number of tokens, causing an enormous resource burden.
Supposing the average degree of an entity is $d$, the number of its neighbors grows exponentially and reaches $d^h$ in the $h$-hop.
While certain sampling strategies such as random walk~\cite{ko2024subgraph} and path pruning~\cite{tan2024paths} have been introduced, a considerable computational load also exists.
As shown in Figure~\ref{fig_intro_fb}, when only sample 20\% 2-hop neighbors in FB15k-237~\cite{DBLP:conf/acl-cvsc/ToutanovaC15} dataset,
the median and mean number of neighbors for entities are about 10 and 107, which requires median and mean tokens of about 300 and 3K, respectively.
When with 30\% sampling, even the median needed tokens reach about 2.5K per entity.
Considering that KG tasks may involve multiple entities, even the most advanced long-context LLMs may face challenges in handling them.
Meanwhile, employing KGs' substructures through sampling could disrupt the holistic modeling of the entire graph, potentially resulting in information loss and sub-optimal performance for downstream tasks.


Another alternate strategy involves integrating continuous KG embedding with LLMs by a learnable adapter~\cite{DBLP:conf/mm/00090GX0C24}, introducing new networks in the framework.
It requires additional precise alignment between the different latent representation spaces of KG embeddings and LLMs.
Considering the above context, we aim to explore the potential to bridge the natural gap between KG structure and natural language and then integrate KGs with LLMs.
Inspired by the early fusion strategy in multimodal LLMs~\cite{DBLP:journals/corr/abs-2405-09818}, the general idea of this study is to learn \emph{compressed} and \emph{discrete} entity codes (\ie, tokens), rather than continuous embeddings, by quantized techniques to represent holistic structural and semantic information of entities in KGs.
They have the same discrete form of natural language,
\eg, the \emph{quantized codes} in Figure~\ref{fig_intro} (b) align the format of language sentences.
Thus, seamlessly integrating KGs with LLMs can be realized by directly inputting the learned codes into LLMs,
merely requiring an expansion of the LLMs' tokenizer vocabulary and eliminating the need for any other framework modifications.

Although several studies have conducted quantized representations on KGs~\cite{DBLP:conf/iclr/0001DWH22,DBLP:conf/aaai/ChenZYZGPC23,DBLP:conf/emnlp/LiWLZM23}, 
they universally employ an unsupervised approach to select anchors to represent entities, failing to the holistic structural and semantic modeling.
In this study, we first introduce a self-supervised quantized representation for KGs, aiming to learn discrete codes for each entity that can reconstruct KG structures and align with semantic texts.
A graph convolutional network (GCN) is used as an encoder to model neighbor structures of KGs, and vector quantization~\cite{van2017neural} is implemented for the KG quantized representation learning.
Further, based on learned entity codes, we construct specific instructions for KG tasks, which can be seamlessly integrated with LLMs, presenting a new paradigm to employ LLMs in KG applications.
In summary, our contributions lie in the following three folds:


$\bullet$\;We propose a self-supervised quantized representation (SSQR) method
that is capable of acquiring both KG structural and semantic knowledge.
To our knowledge, this is the first study for KG quantization learning in a self-supervised manner.

$\bullet$\;We propose the first study that utilizes the derived codes to seamlessly integrate KGs with LLMs, which is achieved by viewing codes as input features and designing KG instruction-following data.
It has extensive potential applications, \eg, KG link prediction and triple classification.

$\bullet$\;From the experiment view, SSQR exhibits superior performance compared to current unsupervised quantized methods and the learned codes are more distinguishable.
Besides, using only 16 codes for each entity, the fine-tuned LLaMA2 and LLaMA3.1 have superior performance on KG link prediction and triple classification tasks.


\section{Quantized Representation for KGs}
\label{sec_quantized}

\begin{figure*}[t]
\centering
\includegraphics[width=0.9\linewidth]{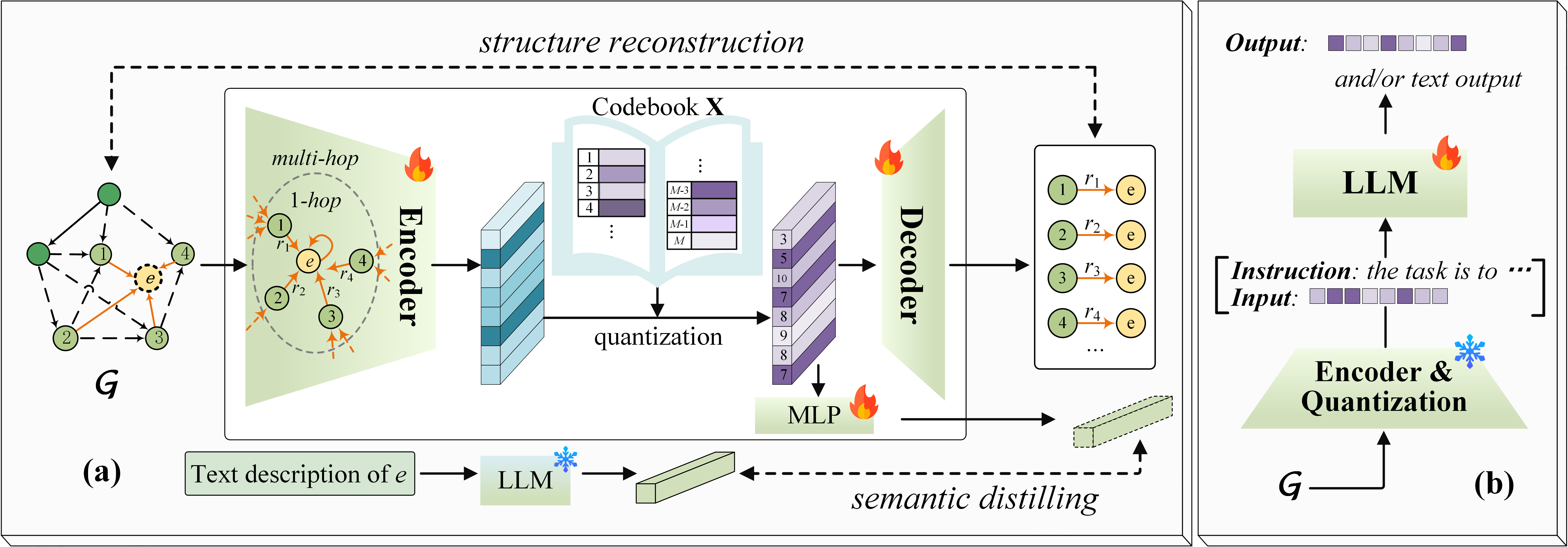}
\setlength{\belowcaptionskip}{-0.3cm}
\caption{
The overall architecture of our study. (a) is for SSQR learning. (b) is for instruction tuning for KG tasks, where the learned quantized representations serve as features.
Icons \smallsnow\;and \smallfire\;represent the status of the module during training, indicating if it is frozen or being updated, respectively.
}
\label{fig_arc}
\end{figure*}

Formally, a KG can be represented as $\mathcal{G}=\{\mathcal{E},\mathcal{R},\mathcal{F}\}$, which is the combination of entities $\mathcal{E}$, relations $\mathcal{R}$, and triples $\mathcal{F}\subseteq \mathcal{E}\times\mathcal{R}\times\mathcal{E}$.
Each triple is in the form of ($h$, $r$, $t$).
For each entity $e$, it has the structural and semantic information, where we utilize the entity neighbors $\mathcal{N}(e)$ and its textual description $T_e$ to describe, respectively.
Although here we only use one-order neighbors $\mathcal{N}(e)$ for demonstration, our model is capable of capturing high-order structure information by multi-layer GCNs.
In the following contents, we will detailedly describe the structural and semantic modeling, as well as the quantized representation for KGs.

\noindent \textbf{Structural Modeling.}
Here, we utilize simple but effective GCNs~\cite{lin2022incorporating} to embed the structural information of KGs, which follows the iterative message-passing strategy to update the entity embeddings from $l$-th layer to ($l$+1)-th:
\begin{equation}
\label{eq_gnn}
  {\mathbf e}_j^{l+1}={\mathbf W}_{1}^{l}{\mathbf e}_j+\!\!\!\!\sum_{(e_i,r)\in \mathcal{N}(e_j)}\!\!\!\!{\mathbf W}_{2}^{l}{\mathbf m}_{e_i,r,e_j}^l,
\end{equation}
where ${\mathbf e}$ is the embedding of the entity and the ${\mathbf W}$ denotes the transformation matrix.
${\mathbf m}$ is the message information of the specific edge.
Here, we follow the composition operation~\cite{DBLP:conf/iclr/VashishthSNT20} for calculation:
\begin{equation}
\label{eq_message}
  {\mathbf m}_{e_i,r,e_j}^l={\mathbf e}_i^l * {\mathbf v}_r^l,
\end{equation}
where ${\mathbf v}$ is the relation embedding and $*$ is element-wise multiplication for two vectors.
Between different layers, relation representation is updated by linear transformation:
${\mathbf v}^{l+1}={\mathbf W}_{rel}^{l}{\mathbf v}^l$.
After $L$ GCN layers, the entity representation ${\mathbf e}^L$ and relation representation ${\mathbf v}^L$ are all obtained.

\noindent \textbf{Quantized Representation.}
Here, we introduce the quantized representation for discrete KG representation.
For its implementation, inspired by VQ-VAE~\cite{van2017neural} and VQ-GAN~\cite{esser2021taming}, we first maintain a discrete cookbook ${\mathbf X}=[{\mathbf x}_1,{\mathbf x}_2,\cdots,{\mathbf x}_M]$, where each ${\mathbf x}_m\in \mathbb{R}^d$ is a learnable vector to represent code $m$.
Using this, a $d$-dimensional vector $\mathbf{e}$ can be quantized by matching the nearest code:
\begin{equation}
\label{eq_quantize}
\textit{Q}(\mathbf{e})=\mathbf{x}_i,\;\text{where}\; i=\underset{m}{\mathop{\arg\min}}\| \mathbf{e}-\mathbf{x}_m \|_2^2,
\end{equation}
where \textit{Q} is quantized function. In this way, each vector can be assigned to only one code, which may lack representation capacity and distinguishability for KG embedding.
So we first transform the learned entity embedding $\mathbf{e}^L$ to multiple times of dimension $d$, \ie, $\textsc{FFN}(\mathbf{e}^L)\in \mathbb{R}^{N\times d}$.
In this way, each entity can be assigned to a code sequence with the length of $N$.
Thus, each entity can be represented to $[q_1,q_2,\cdots,q_N]$ by Eq. (\ref{eq_quantize}), where $q_n$ is the code index in the codebook. The quantized representation vector can be:
\begin{equation}
\textbf{q}_e=\textbf{W}_q\textit{Q}(\mathbf{e}^L),\textit{Q}(\mathbf{e}^L)=[\mathbf{x}_{q_1},\mathbf{x}_{q_2}\cdots\mathbf{x}_{q_N}].
\end{equation}
Based on this, the whole model can be optimized in an end-to-end manner by the straight-through gradient estimator~\cite{van2017neural}:
\begin{equation}
    \mathcal{L}_{q} = \big\| \text{sg}[\mathbf{e}^L] - \textbf{q}_e \big\|_2^2 + \beta \big\| \mathbf{e}^L - \text{sg}[\textbf{q}_e] \big\|_2^2,
\end{equation}
where \text{sg} stands for \emph{stop gradient}, which is characterized by its identity function during forward computation and has zero partial derivatives for backpropagation.
$\| \text{sg}[\mathbf{e}^L] - \textbf{q}_e \big\|_2^2$ is codebook loss assuring the codes are close to encoder’s outputs and $\| \mathbf{e}^L - \text{sg}[\textbf{q}_e] \big\|_2^2$ is commit loss encouraging the encoder generating outputs close to codes.
$\beta$ is a hyper-parameter to trade off the two terms.

\noindent \textbf{Structure Reconstruction.}
To inject the holistic structure information into the quantized representations,
we hope the learned entity codes can reconstruct KG structures.
But directly predicting adjacency matrix as is done in research for homogeneous graphs~\cite{DBLP:conf/iclr/0006TXLHQZ0ZL24} is inappropriate,
because KGs' structures are more heterogeneous and sparse. Thus, based on quantized embeddings, we verify the validity of each triplet ($h$, $r$, $t$) and implicitly reflect the holistic KG, where ConvE~\cite{dettmers2018convolutional} is implemented:
\begin{equation}
\label{eq_hrt}
  s(h,r,t)=\big[\textsl{Flat}(\textsl{Conv}(\bar{{\mathbf q}}_h\|\bar{{\mathbf v}}_r^L))\big]^\top{\mathbf W}_{c}{\mathbf q}_t.
\end{equation}
\textsl{Flat} and \textsl{Conv} are the flatten and 2D convolution operations, respectively.
$\bar{\mathbf q}$ and $\bar{\mathbf v}$ are transformation matrices for embeddings ${\mathbf q}$ and ${\mathbf v}$.
The final score of triple ($h$, $r$, $t$) can be regularized by the sigmoid function $\sigma$: $\tilde{y}=\sigma(s(h,r,t))$.
Finally, compared with actual label $y$, the structure modeling can be learned by binary cross-entropy loss:
\begin{equation}
\label{eq_str}
  \mathcal{L}_{st}\!=\!-\frac{1}{|\mathcal{F}|}\!\sum_i [y_i\log \tilde{y}_i+(1-y_i)\log (1\!-\!\tilde{y}_i)].
\end{equation}

\noindent \textbf{Semantic Distilling.}
For semantic modeling, our goal is to ensure that the learned codes for each entity can imply the information of its corresponding text descriptions.
Considering the substantial success of LLMs, we introduce a simple yet effective distilling strategy to learn from them.
Specifically, we first obtain text embeddings of KG entities by LLMs:
${\mathbf t}_{e}=\textsl{LLM}(T_e)$.
Here, we utilize the \emph{text-embedding-3-large} as the LLM by OpenAI API~\footnote{https://platform.openai.com/docs/guides/embeddings} considering its strong ability for text embeddings.
It embeds each text sequence into a 3072-dimension vector.
Based on this, we make the model have the ability to align its semantic embedding through the learned quantized output, where the loss of mean square error is utilized:
\begin{equation}
\label{eq_mse}
  \mathcal{L}_{se}=-\frac{1}{|\mathcal{E}|}\sum_i \big\|{\mathbf W}_{s}{\mathbf q}_{e_i}-{\mathbf t}_{e_i}\big\|_2^2.
\end{equation}
In this way, we distil the semantic knowledge from the LLMs to our discrete codes of GCN outputs.

On the whole, the entire quantized representation model can be updated by the combination of quantized, structural, and semantic loss:
\begin{equation}
  \mathcal{L}=\mathcal{L}_{q}+\mathcal{L}_{st}+\mathcal{L}_{se}.
\end{equation}

\section{Tuning LLMs with SSQR}
\label{sec_llms}

Employing the quantized representation, each entity in KGs can be illustrated by codes of length 
$N$.
This can be perceived as the same form of natural language, thereby facilitating its seamless integration with LLMs.
Every learned code can serve as a new token, necessitating only an expansion of the token vocabulary within the LLM's tokenizer.

This paradigm can be applied to various KG tasks, by constructing corresponding instruction data, where the learned entity codes could act as features.
For example, the KG link prediction task can be done using the instruction form as shown in Table~\ref{tab_instruction}.
Specifically, the code sequence of entity $e$ can be \textsl{Code}($e$): ``[CODE$q_1$] [CODE$q_2$] $\cdots$ [CODE$q_N$]''.
For each query ($h$, $r$, ?), we provide the codes \textsl{Code}($h$) for query head $h$.
Besides, we give several answer candidates along with their associated codes for LLM ranking.
Candidates can be obtained by conventional KG embedding models, \eg, TransE~\cite{DBLP:conf/nips/BordesUGWY13} and CompGCN~\cite{DBLP:conf/iclr/VashishthSNT20}.
The goal is to predict the actual ranking list of candidates using their discrete codes.
The detailed instructions for triple classification are described in Section~\ref{app_details} of the Appendix.
For the LLM fine-tuning,
the next token prediction is carried out based on the instruction $\mathcal{I}$ and previously generated text tokens: 
\begin{equation}
  \mathcal{L}_{llm}=-\sum_{n=1}^N \log\big(x_n|x_{<n},\mathcal{I}\big).
\end{equation}

\begin{table}[]
    \centering
    \begin{tabular}{c}
                \begin{tcolorbox}[colback=gray!10,
                      colframe=black,
                      width=7.cm,
                      boxrule=0.8pt,
                      arc=1mm, auto outer arc,
                      left = 1mm, 
                        right = 1mm,
                        top = 1mm,
                        bottom = 1mm,
                     ]
                \footnotesize{
                \textbf{Instruction:} This is a knowledge graph completion task, which needs to predict the tail entity for an incomplete query triplet.\\
                \textbf{Input:} The query triplet is (\emph{h}, \emph{r}, ?).\\The quantized representation of entity \emph{h} is: \blue{[Code(\emph{h})]}\\
                The answer candidates and corresponding quantized representations are as follows:\\
                \emph{entity 1},\;\blue{[Code(\emph{entity 1})]}\\
                $\cdots$\\
                \emph{entity 20},\;\blue{[Code(\emph{entity 20})]}\\
                Please generate quantized representations of the top-3 potential answers, ranked from highest to lowest:\\
                \textbf{Output:} 1.\;\blue{[Code(\emph{candidate 1})]}\\
                2.\;\blue{[Code(\emph{candidate 2})]}\\
                3.\;\blue{[Code(\emph{candidate 3})]}
                }
                \end{tcolorbox} \\
    \end{tabular}
    \caption{Instruction format for link prediction, where learned codes serve as entity features to help ranking.}
    \label{tab_instruction}
\end{table}

\begin{figure*}[t]
\centering
\hspace{-0.25cm}
\subfloat[Original text embedding.]{
\includegraphics[width=0.28\linewidth]{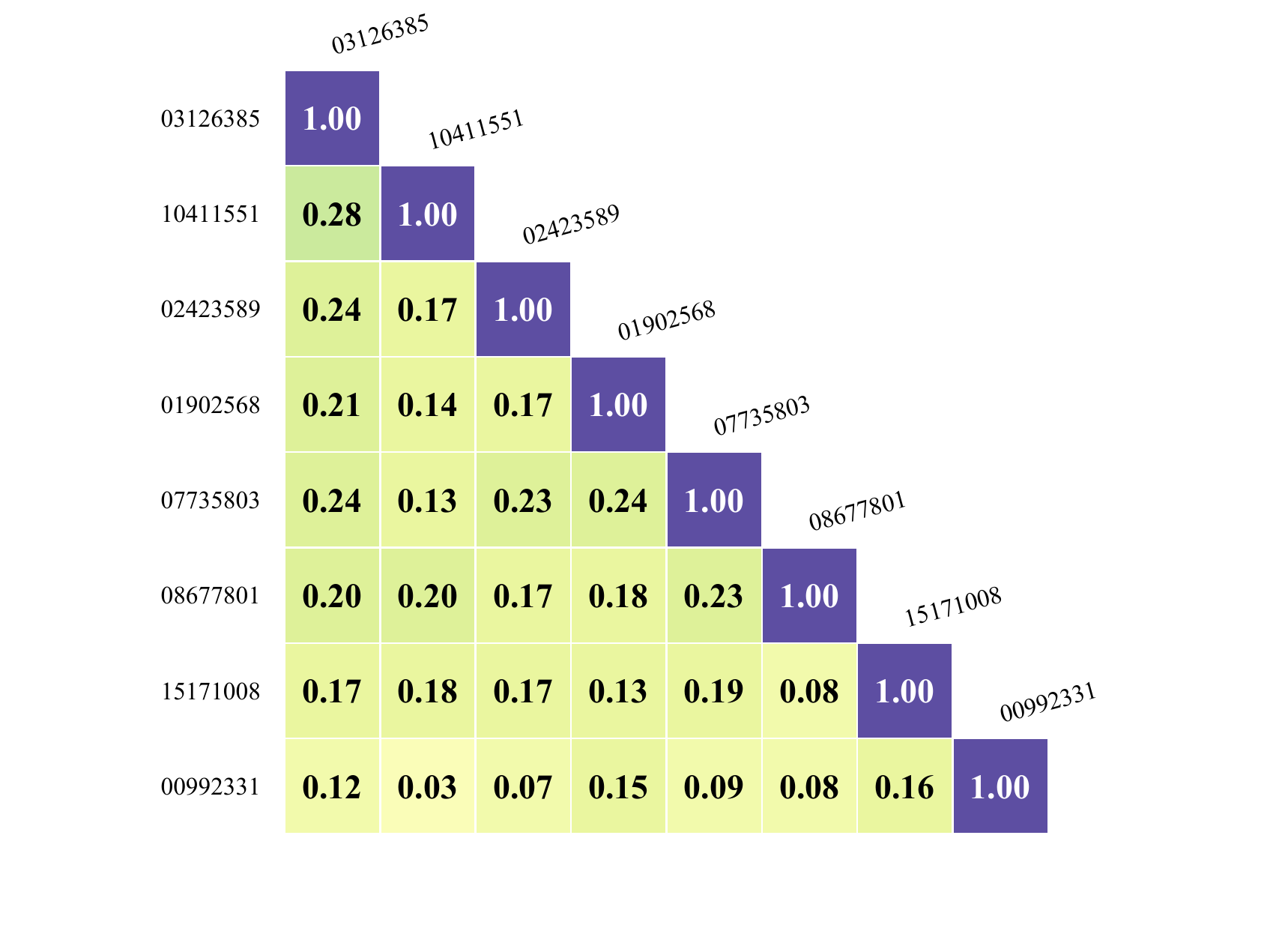}}
\hspace{-0.85cm}
\subfloat[SSQR.]{
\includegraphics[width=0.28\linewidth]{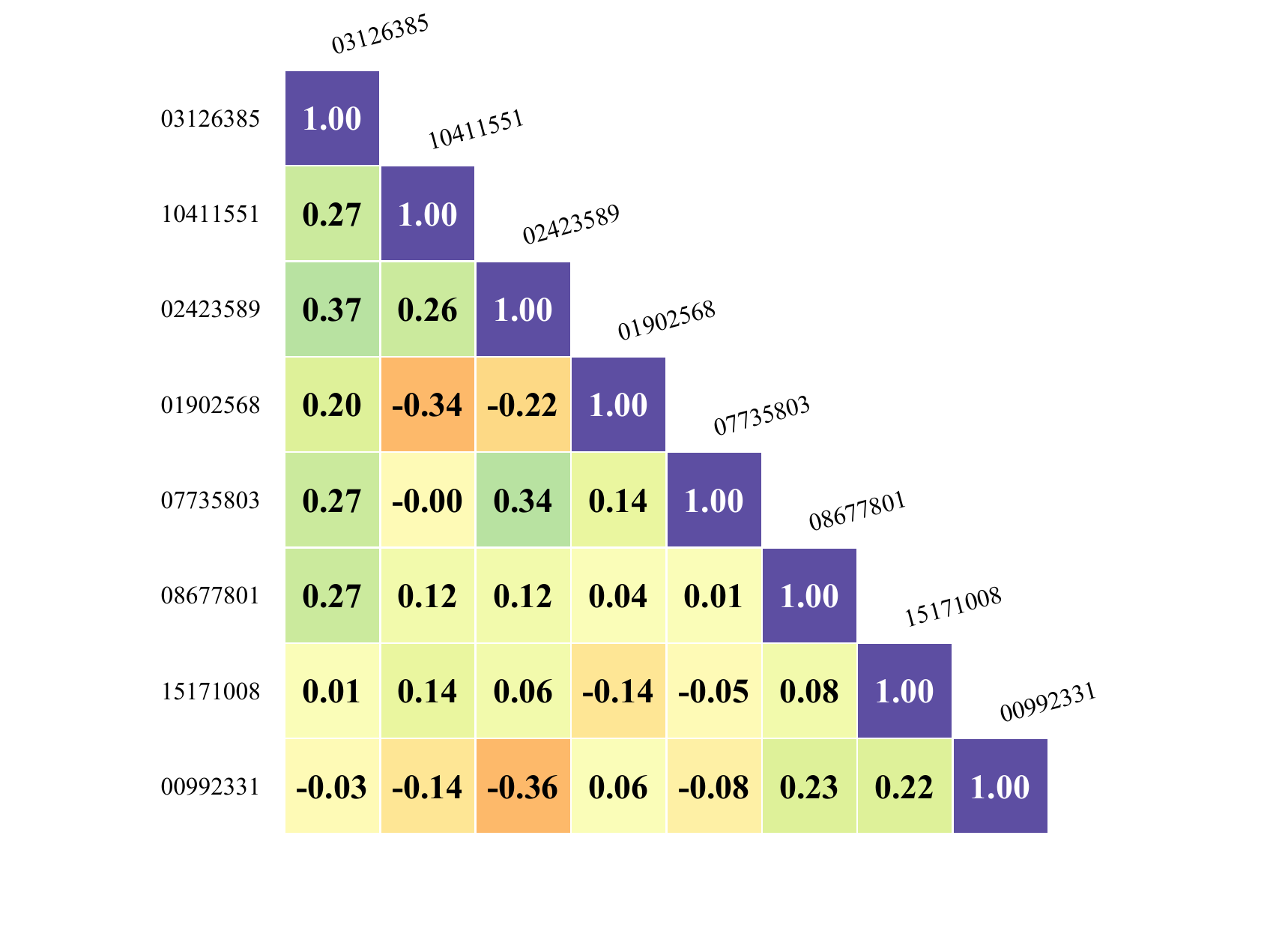}}
\hspace{-0.85cm}
\subfloat[SSQR w/o GCN.]{
\includegraphics[width=0.28\linewidth]{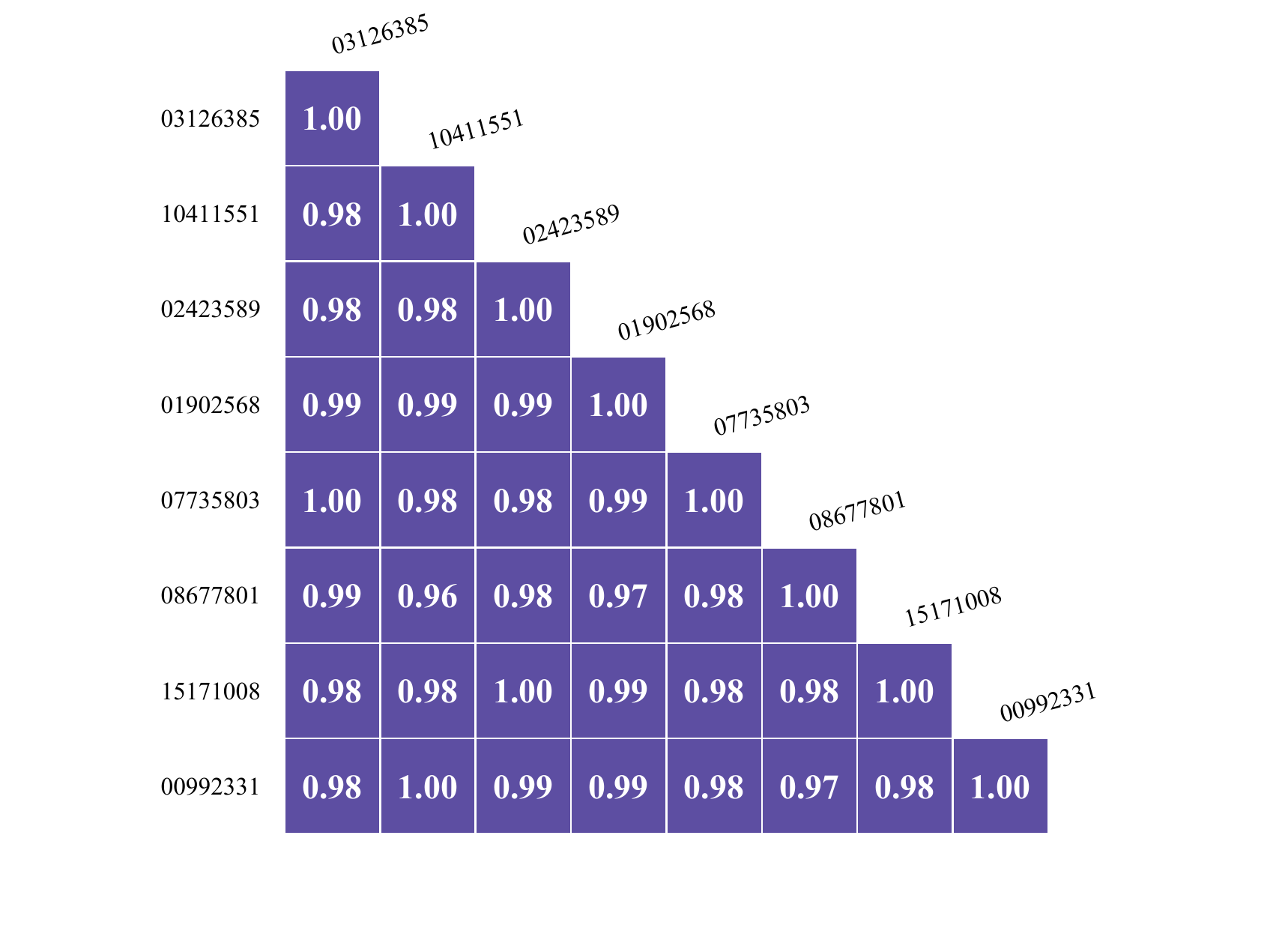}}
\hspace{-0.85cm}
\subfloat[SSQR w/o semantics.]{
\includegraphics[width=0.28\linewidth]{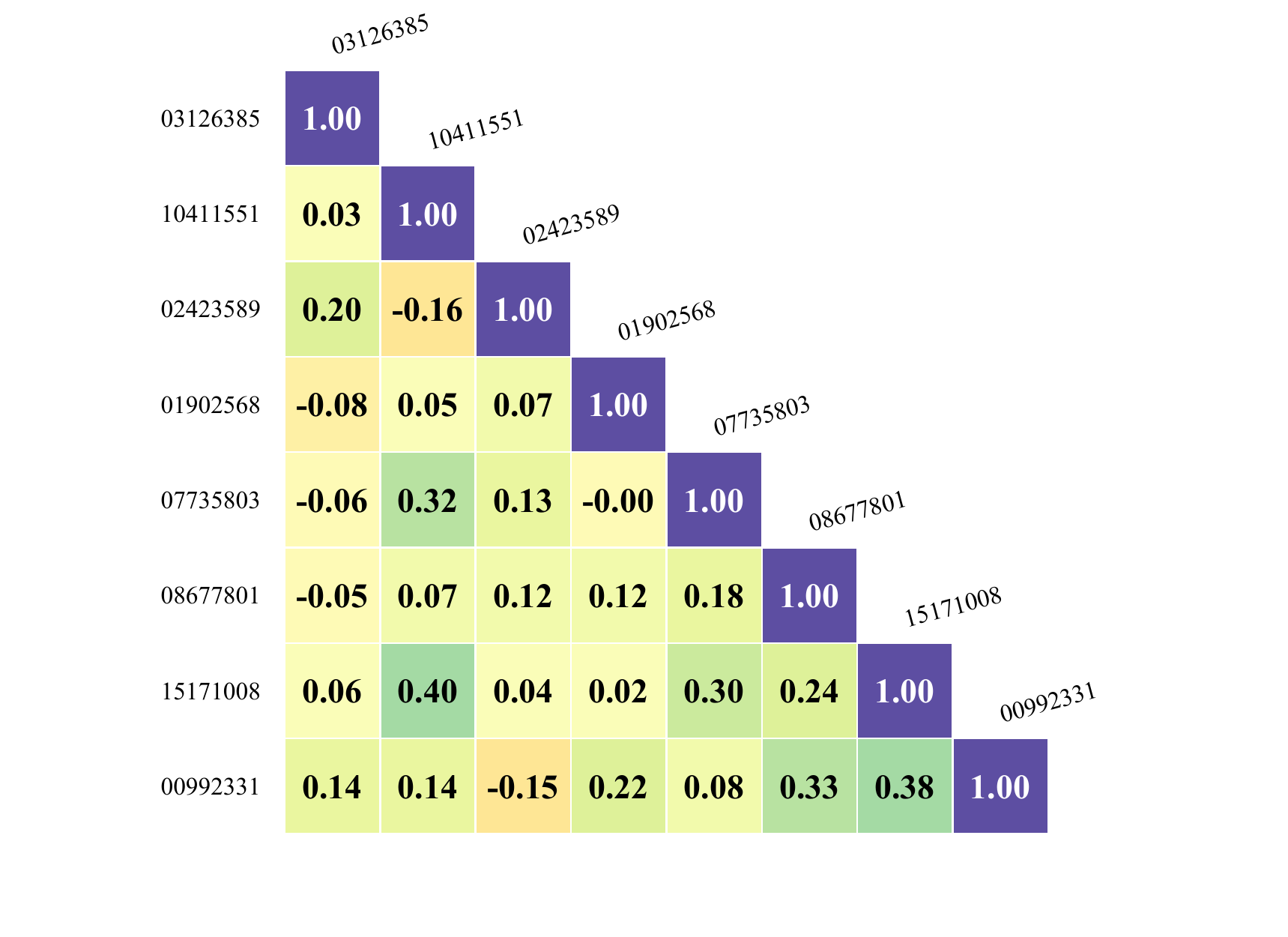}}
\setlength{\abovecaptionskip}{0.1cm}
\setlength{\belowcaptionskip}{-0.3cm}
\caption{The cosine similarity of quantized representations on the WN18RR dataset (sampled 8 entities).}
\label{fig_simwn}
\end{figure*}

\section{Experiments and Analysis}
\label{sec_exp}

To verify the effectiveness of the proposed SSQR and its ability to integrate with LLMs,
We carry out experiments on the KG link prediction and triple classification tasks, where the popular datasets WN18RR~\cite{dettmers2018convolutional} and FB15k-237~\cite{DBLP:conf/acl-cvsc/ToutanovaC15} as well as FB15k-237N~\cite{DBLP:conf/acl/LvL00LLLZ22} are utilized.
For SSQR, a 2-layer GCN is utilized as the encoder.
$\beta$ is set to 0.25 in the experiment.
The embedding dimension is set to 200 as default.
The number of codebook $M$ and codes for each entity $N$ is set to 2048 and 32.
The maximum number of training epochs is 800.
For LLM fine-tuning, LLaMA2 (7B) and LLaMA3.1 (8B) are utilized using $M$ as 2048 and $N$ as 16 for computation efficiency.
The query for head prediction (?, $r$, $t$) is transformed to the tail prediction by adding reverse relation of $r$.
The mean reciprocal rank (MRR) and Hits@$k$ values are set as evaluation metrics for model performance.
Moreover, the triple classification task employs accuracy, precision, recall and F1-score as metrics.
More detailed settings are shown in the Appendix.


\begin{table}[t!]
\setlength{\abovecaptionskip}{0.1cm}
\caption{The results of baselines are from~\citet{DBLP:conf/emnlp/LiWLZM23}.
$\dagger$ means the improvement of SSQR compared to the best performance in each metric.
$\ddagger$ means the ablation results compared to the results of SSQR.}
\centering
\resizebox{0.48\textwidth}{!}{
\begin{tabular}{lcccc}
\toprule
\multicolumn{1}{c}{\multirow{2}{*}{Model}}& \multicolumn{2}{c}{WN18RR}   & \multicolumn{2}{c}{FB15k-237}  \\
\cmidrule(r){2-3} \cmidrule(r){4-5}
\multicolumn{1}{c}{}& MRR & Hits@10 & MRR  & Hits@10 \\
\midrule
NodePiece &0.403 &0.515 &0.256 &0.420\\
$\;\;$+RandomEQ& 0.425& 0.522 &0.263& 0.425\\
\midrule
EARL &0.440 &0.527 &\underline{0.310}& 0.501\\
$\;\;$+RandomEQ &\underline{0.442} &\underline{0.536} & 0.308& \underline{0.502}\\
\midrule
SSQR&\textbf{0.483} &\textbf{0.578} &\textbf{0.361}& \textbf{0.545}\\
$\quad\;\;\Delta$ ($\uparrow$)$^\dagger$&{\cellcolor{gray!20}9.28\%}	&{\cellcolor{gray!20}7.84\%}	 &{\cellcolor{gray!20}16.45\%}& {\cellcolor{gray!20}8.57\%}\\
\midrule
$\quad\;\;$w/o GCN&0.479 &0.577 &0.309& 0.482\\
$\quad\;\;\Delta$ ($\downarrow$)$^\ddagger$&0.83\%&	0.17\%& {\cellcolor{gray!20}14.40\%}& {\cellcolor{gray!20}11.56\%} \\
$\quad\;\;$w/o sem&0.447 &0.521 &0.347& 0.528\\
$\quad\;\;\Delta$ ($\downarrow$)$^\ddagger$&{\cellcolor{gray!20}7.45\%} &{\cellcolor{gray!20}9.86\%}&3.88\%& 3.12\%\\
\bottomrule
\end{tabular}
}
\label{tab_qrresults}
\end{table}

\subsection{SSQR Results}
We compare the performance of our SSQR with three unsupervised methods, \ie, NodePiece~\cite{DBLP:conf/iclr/0001DWH22}, EARL~\cite{DBLP:conf/aaai/ChenZYZGPC23}, and random entity quantization (RandomEQ for short)~\cite{DBLP:conf/emnlp/LiWLZM23}, for KG quantized representations. The results are given in Table~\ref{tab_qrresults}.

As can be observed, SSQR achieves significant performance improvement against baselines, which has 9.28\% and 7.84\% improvements compared with the previous optimal performance on the WN18RR dataset.
When at the FB15k-237 dataset, the improvements are even better, \ie, 16.45\% and 8.57\%. Although these unsupervised methods are simple and efficient for implementation, they fail to capture the structures of KGs. In contrast, our proposed self-supervised strategies would provide an effective way for quantized representations for KG structure learning.

\subsection{SSQR Result Analysis}

\noindent \textbf{Ablation Studies.}
We carry out the ablation studies to verify the effectiveness of each module in SSQR as the bottom part of Table~\ref{tab_qrresults}.
Generally, the performance of link prediction degrades when GCN or semantic distilling is removed, but the extent of degradation varies across different datasets.
It can be seen that the GCN encoder is more important for the FB15k-237 dataset (14.40\% and 11.56\% decline), while semantic information has more impact on WN18RR (7.45\% and 9.86\%).
This may be due to the fact that FB15k-237 contains a rich KG structure which requires GCN to capture, while the semantic text is more important for WN18RR to make up for the defects caused by the lack of rich structural information.

\noindent \textbf{Relevance among Entity Codes.}
We also calculate the cosine similarity of quantized representation in Figure~\ref{fig_simwn}, including the original text embedding, SSQR, SSQR w/o GCN, and SSQR w/o semantics.
When using only text embeddings, the similarities are all small positive values.
SSQR w/o GCN has similarities that are all close to 1.
These phenomena indicate that entity representations are in a small corner of the space (\ie, \emph{anisotropic}), where the representation space is not fully utilized for efficient representation.
SSQR solves this problem to a certain extent, with a greater range and variety of similarities.
Removing semantic information would diminish that advantage.

\begin{figure}[t]
\centering

\hspace{-0.3cm}
\subfloat[WN18RR dataset.]{
\includegraphics[width=0.51\linewidth]{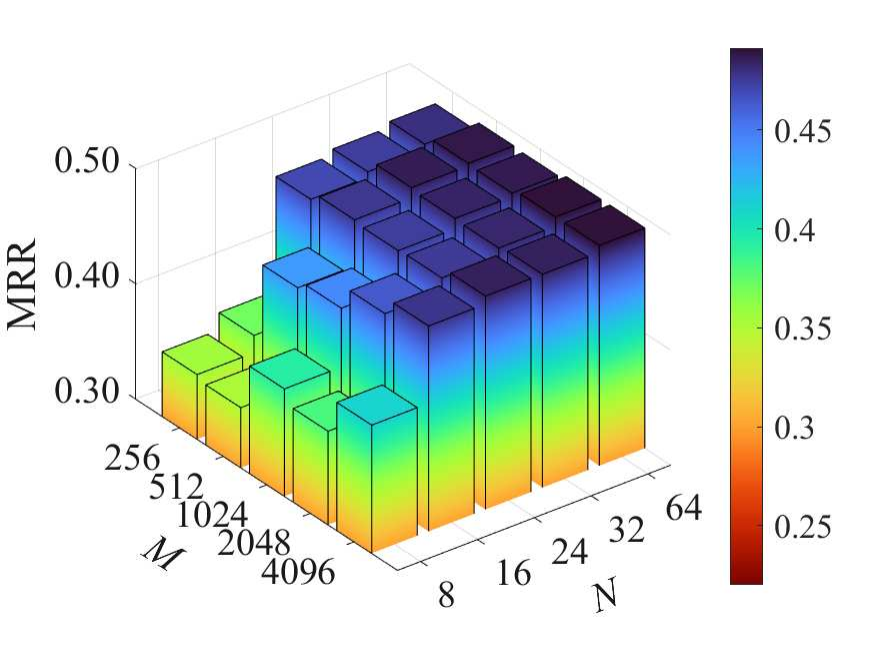}}
\hspace{-0.3cm}
\subfloat[FB15k-237 dataset.]{
\includegraphics[width=0.51\linewidth]{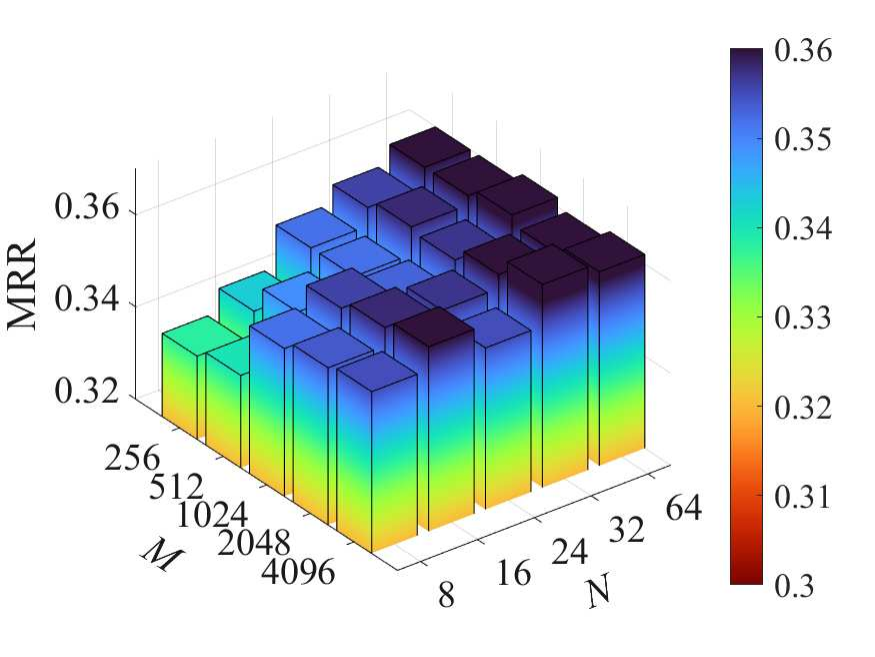}}
\vspace{-0.3cm}
\setlength{\belowcaptionskip}{-0.3cm}
\caption{The effects of codebook length ($M$) and sequence length ($N$) for each entity.
}
\label{fig_mn}
\end{figure}

\noindent \textbf{Impacts of $M$ and $N$.}
The number of codebooks and sequence lengths for codes, \ie, $M$ and $N$, are vital hyper-parameters for SSQR.
We explore their impacts in Figure~\ref{fig_mn}.
Generally, larger $M$ and $N$ would lead to better performance as they increase the modeling ability of SSQR.
In the WN8RR dataset, $N$ has a greater impact on $M$, indicating the necessity of a large $N$ for holistic and distinguishable representations.
It may be caused by the sparser structure and more entities in the WN18RR.

\begin{figure}[t]
\centering

\hspace{-0.45cm}
\subfloat[WN18RR dataset.]{
\includegraphics[width=0.535\linewidth]{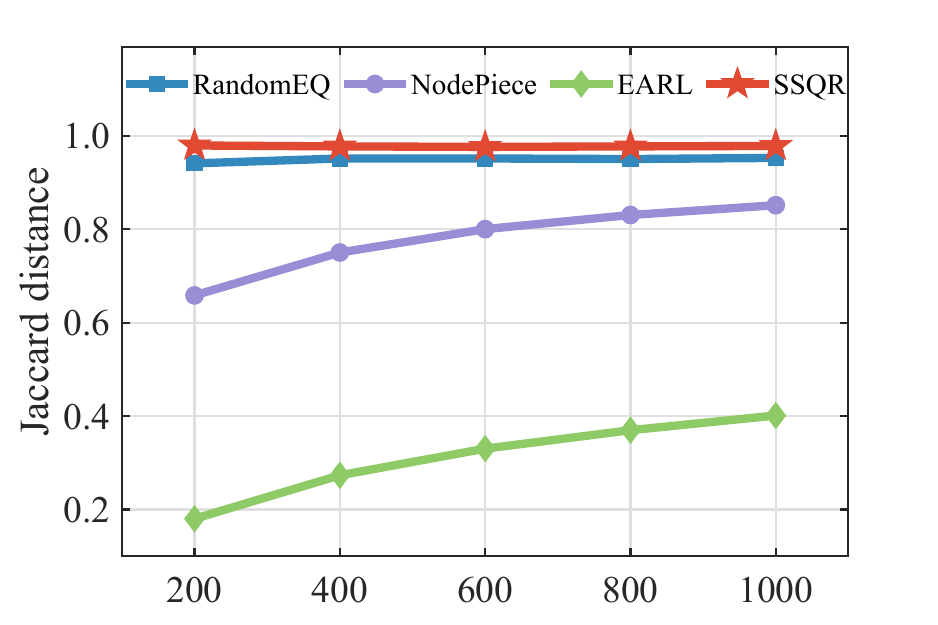}}
\hspace{-0.5cm}
\subfloat[FB15k-237 dataset.]{
\includegraphics[width=0.535\linewidth]{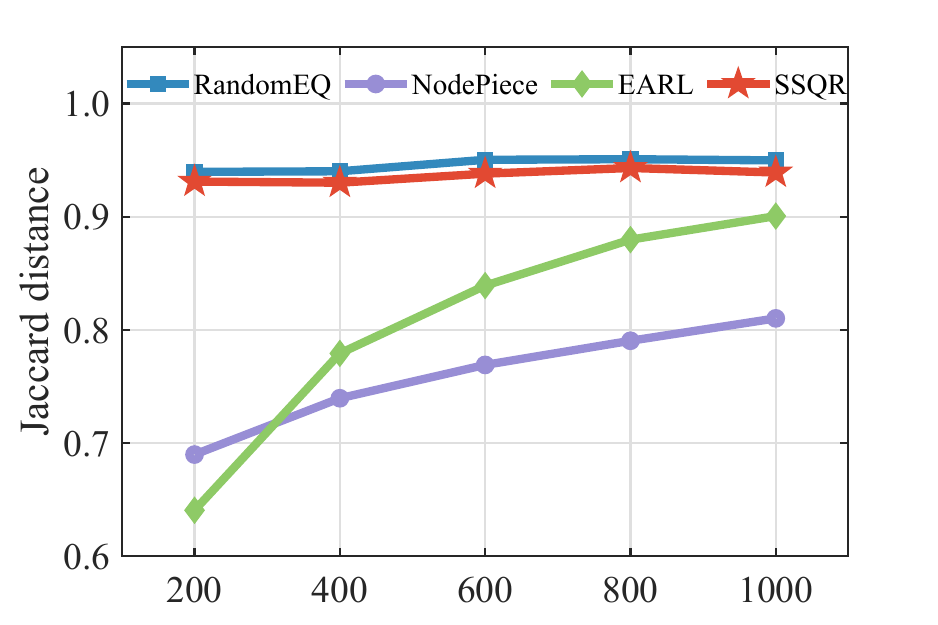}}
\setlength{\abovecaptionskip}{0.1cm}
\setlength{\belowcaptionskip}{-0.3cm}
\caption{The mean Jaccard distance between codes of a specific entity and its $k$ nearest ones.
}
\label{fig_jaccard}
\end{figure}

\noindent \textbf{Distinguishability of SSQR.}
Following RandomEQ, we calculate the general entropy and Jaccard distance to show codes-level and codewords-level distinguishability, respectively.
For general entropy, SSQR has 16.76 and 15.27 on WN18RR and FB15k-237 datasets, similar to RandomEQ (16.75/15.27) and higher than that of NodePiece (15.94/15.26) and EARL (8.20/14.50).
It shows that our method has more diverse entity codes and better entity differentiation ability than other quantization methods.
The Jaccard distances of each model are shown in Figure~\ref{fig_jaccard}.
RandomEQ and SSQR have high values that are far better than those of NodePiece and EARL.
RandomEQ is superior on the FB15k-237 dataset but SSQR performs better on the WN18RR dataset.
In summary, SSQR exhibits a robust capacity to distinguish different entities and effectively represent KGs.

\subsection{Quantized Representations with LLMs}

\paragraph{Link Prediction.}
For fine-tuning, we utilize the pre-trained AdaProp~\cite{zhang2023adaprop} to generate 20 candidates for each query as it has strong and balanced performance on most KG tasks.
For comparison, we selected the current advanced embedding models, like TransE~\cite{DBLP:conf/nips/BordesUGWY13}, CompGCN~\cite{DBLP:conf/iclr/VashishthSNT20}, AdaProp~\cite{zhang2023adaprop}, MA-GNN~\cite{DBLP:conf/acl/XuBL23}, TCRA~\cite{DBLP:conf/acl/Guo0LXN24}, and DiffusionE~\cite{DBLP:conf/kdd/CaoLWL24}.
Besides, we include five advanced LLM-based methods for more direct comparison, including KICGPT~\cite{DBLP:conf/emnlp/WeiH0K23}, CSProm-KG-CD~\cite{DBLP:conf/eacl/LiTCL24}, ARR~\cite{chen2024new}, KG-FIT~\cite{DBLP:journals/corr/abs-2405-16412}, and MKGL~\cite{DBLP:journals/corr/abs-2410-07526}.

The results of link prediction are shown in Table~\ref{tab_llmresult}.
It can be observed that SSQR with LLaMA2 or LLaMA3.1 is obviously superior in KG link prediction against general embedding methods.
Compared with the previous state-of-the-art MA-GNN, SSQR with LLaMA2 achieves about 4.60\%, 8.09\%, 4.39\%, -0.88\% and	18.47\%, 32.62\%, 18.31\%, 4.92\% improvement in two datasets, respectively.

\begin{table*}[t!]
\setlength{\abovecaptionskip}{0.1cm}
\caption{The experiment results of general embedding methods and LLM-based methods for KG link prediction.
}
\centering
\resizebox{0.99\textwidth}{!}{
\begin{tabular}{lcccccccc}
\toprule
\multicolumn{1}{c}{\multirow{2}{*}{\textbf{Model}}}& \multicolumn{4}{c}{\textbf{WN18RR}}   & \multicolumn{4}{c}{\textbf{FB15k237}}  \\
\cmidrule(r){2-5} \cmidrule(r){6-9}
\multicolumn{1}{c}{}& \textbf{MRR}  & \textbf{Hits@1} & \textbf{Hits@3} & \textbf{Hits@10} & \textbf{MRR}  & \textbf{Hits@1} & \textbf{Hits@3} & \textbf{Hits@10} \\
\midrule
\multicolumn{9}{c}{\cellcolor{gray!50}General Embedding Methods}\\
TransE~\cite{DBLP:conf/nips/BordesUGWY13}&0.223& 0.014 &0.401 &0.529&  0.330& 0.231& 0.369 &0.528 \\
CompGCN~\cite{DBLP:conf/iclr/VashishthSNT20}&0.479 &0.443 &0.494 &0.546& 0.355& 0.264& 0.390 &0.535\\
AdaProp~\cite{zhang2023adaprop}&0.562 &0.499& --&0.671& \underline{0.417}& \underline{0.331}&--& \underline{0.585}\\
MA-GNN~\cite{DBLP:conf/acl/XuBL23}&\underline{0.565}&  \underline{0.507}& \underline{0.592}& \underline{0.679}& 0.379 &0.282& \underline{0.415}& 0.569\\
TCRA~\cite{DBLP:conf/acl/Guo0LXN24}& 0.496& 0.457& 0.511& 0.574& 0.367& 0.275& 0.403& 0.554\\
DiffusionE~\cite{DBLP:conf/kdd/CaoLWL24}& 0.557& 0.504& --& 0.658& 0.376& 0.294& -- &0.539\\
\midrule
\multicolumn{9}{c}{\cellcolor{gray!50}LLM-based Methods}\\
KICGPT~\cite{DBLP:conf/emnlp/WeiH0K23} &0.549& 0.474& 0.585 &0.641& 0.412&\underline{0.327}& 0.448 &0.554\\
CSProm-KG-CD~\cite{DBLP:conf/eacl/LiTCL24} &\underline{0.559}& \underline{0.508}& 0.578 &0.660 &--&--&--&--\\
ARR~\cite{chen2024new}&0.521&--&\underline{0.607}&--&0.398 &--&0.436&--\\
KG-FIT~\cite{DBLP:journals/corr/abs-2405-16412}&0.553& 0.488& 0.595& \textbf{0.695}& 0.362 &0.275& \underline{0.485}& 0.572\\
MKGL~\cite{DBLP:journals/corr/abs-2410-07526}&0.552& 0.500& 0.577& 0.656& \underline{0.415}& 0.325& 0.454& \underline{0.591}\\
\midrule
SSQR-LLaMA2 &\textbf{0.591} &\textbf{0.548} &\textbf{0.618} &\underline{0.673}&\textbf{0.449} & \textbf{0.374}  &\textbf{0.491} &\textbf{0.597}\\
SSQR-LLaMA3.1&\textbf{0.598} &\textbf{0.559} &\textbf{0.618} &\underline{0.675}&\textbf{0.459} & \textbf{0.393}& \textbf{0.491}& \textbf{0.597}\\
\bottomrule
\end{tabular}
}
\label{tab_llmresult}
\end{table*}

Compared with LLM-based methods, SSQR-LLaMA2 also shows competitive performance.
It is better than KICGPT, CSProm-KG-CD, and ChatGPT.
Even KICGPT achieves good results on the FB15k-237 dataset, it can also be raised by 8.98\%, 14.37\%, 9.60\%, and 7.76\%.
For the KG-FIT (HAKE), it also has 6.87\%, 12.30\%,	3.87\%, and	-3.16\% improvements on the WN18RR dataset.
Although there is a slight deficiency in terms of Hits@10, improvements on other metrics are high.
Meanwhile, SSQR-LLaMA3.1 is better than SSQR-LLaMA2, demonstrating that learned quantized representations can be used for a more powerful LLM to get better performance.
From all the results, our methods generally achieve a greater improvement in the Hits@1 metric, which is caused by the candidate selection and ranking strategies we used in LLM fine-tuning.
The candidate selection model may have limited ability, but our method has a strong ability to select the correct answer from all candidates.
This demonstrates that our method has good scalability and can be further improved with more accurate candidate selection models.

\paragraph{Triple Classification.}
Beyond the link prediction task, we conduct experiments on triple classification on the FB15k-237N dataset.
The results are shown in Table~\ref{tab_llmtc}, where our method outperforms general embedding methods and other LLM-based baselines.
For the advanced KoPA~\cite{DBLP:conf/mm/00090GX0C24} model, the performance in the F1-score metric is comparable to that of SSQR.
However, the accuracies of SSQR show a significant improvement, \ie, 0.794/0.798 vs. 0.777, demonstrating the effectiveness of integrating SSQR with LLMs.

\begin{table}[t!]
\setlength\tabcolsep{3pt} 
\setlength{\abovecaptionskip}{0.1cm}
\caption{The experiment results of the triple classification on FB15k-237N dataset. The results of baselines are taken from~\citet{DBLP:conf/mm/00090GX0C24}.
}
\centering
\resizebox{0.49\textwidth}{!}{
\begin{tabular}{lcccc}
\toprule
\multicolumn{1}{c}{\textbf{Model}}& {\cellcolor{gray!20}\textbf{Acc}}  & \textbf{P}&\textbf{R}&{\cellcolor{gray!20}\textbf{F1}} \\
\midrule
TransE~\cite{DBLP:conf/nips/BordesUGWY13}&{\cellcolor{gray!20}0.697} &0.708 &0.671 &{\cellcolor{gray!20}0.689}\\
DistMult~\cite{DBLP:journals/corr/YangYHGD14a}&{\cellcolor{gray!20}0.587} &0.590& 0.568 &{\cellcolor{gray!20}0.579}\\
RotatE~\cite{DBLP:conf/iclr/SunDNT19}&{\cellcolor{gray!20}0.684} &0.692& 0.664 &{\cellcolor{gray!20}0.678}\\
\midrule
Alpaca$_{\textsl{zero-shot}}$&{\cellcolor{gray!20}0.561}& 0.533& 0.974& {\cellcolor{gray!20}0.689}\\
GPT-3.5$_{\textsl{zero-shot}}$&{\cellcolor{gray!20}0.602}& 0.866& 0.240& {\cellcolor{gray!20}0.376}\\
KG-LLaMA~\cite{DBLP:journals/corr/abs-2308-13916}&{\cellcolor{gray!20}0.748} &0.674& 0.962& {\cellcolor{gray!20}\underline{0.793}}\\
KG-Alpaca~\cite{DBLP:journals/corr/abs-2308-13916}&{\cellcolor{gray!20}0.699}& 0.627& 0.983& {\cellcolor{gray!20}0.766}\\
KoPA~\cite{DBLP:conf/mm/00090GX0C24}&{\cellcolor{gray!20}\underline{0.777}}& 0.708& 0.941 &{\cellcolor{gray!20}\textbf{0.808}}\\
\midrule
SSQR-LLaMA2 &{\cellcolor{gray!20}\textbf{0.794}}&	0.757&	0.867&	{\cellcolor{gray!20}\textbf{0.808}}\\
$\quad\quad\quad\quad$w/o SSQR&{\cellcolor{gray!20}0.754}&	0.699&	0.891&	{\cellcolor{gray!20}0.783}\\
$\quad\quad\quad\quad\Delta$&{\cellcolor{gray!20}-5.13}\%&-7.71\%	&+2.85\%&{\cellcolor{gray!20}-3.07\%}\\
\midrule
SSQR-LLaMA3.1 &{\cellcolor{gray!20}\textbf{0.798}}&	0.759&	0.872&	{\cellcolor{gray!20}\textbf{0.811}}\\
$\quad\quad\quad\quad$w/o SSQR&{\cellcolor{gray!20}0.767}&	0.711&	0.901&	{\cellcolor{gray!20}0.795}\\
$\quad\quad\quad\quad\Delta$&{\cellcolor{gray!20}-3.77\%}&	-6.34\%&	+3.41\%&	{\cellcolor{gray!20}-2.03\%}\\
\bottomrule
\end{tabular}
}
\label{tab_llmtc}
\end{table}

\subsection{Insights of LLM Fune-tuning}

\noindent \textbf{Ablation Studies.}
We carry out ablation studies to verify the effectiveness of quantized representations for LLM tuning. The results are shown in Table~\ref{tab_abl_llm} and the bottom part of Table~\ref{tab_llmtc}, where \emph{w/o SSQR} means only utilizing the entity's name for fine-tuning and removing learned entity codes.
For the link prediction task, there is a large performance drop, especially in the MRR, Hits@1, and Hits@3 metrics. A similar pattern is also present in the triple classification task. We observe that when under the \emph{w/o SSQR} setting, LLMs have overfitting issues, where their performance on training sets is very high but fails to generalize to valid and test sets.
This demonstrates that the learned discrete codes are distinguishable and representative for different entities, thereby allowing their utilization as features to assist KG tasks in LLMs.

\begin{table}[t!]
\setlength{\abovecaptionskip}{0.1cm}
\caption{The ablation results for the link prediction task.}
\centering
\resizebox{0.48\textwidth}{!}{
\begin{tabular}{lcccc}
\toprule
\multicolumn{1}{c}{Model}& MRR & Hits@1 & Hits@3  & Hits@10 \\
\midrule
\multicolumn{5}{c}{\cellcolor{gray!50}WN18RR}\\
SSQR-LLaMA2 &0.591& 0.548& 0.618 &0.673 \\
$\quad$w/o SSQR &0.541	&0.495	&0.603	&0.668	\\
$\quad\quad\;\;\;\Delta$ ($\downarrow$)&{\cellcolor{red!20}8.46\%}&	{\cellcolor{red!20}9.67\%}&	2.43\%&	0.74\%\\
\midrule
\multicolumn{5}{c}{\cellcolor{gray!50}FB15k-237}\\
SSQR-LLaMA2 &0.449& 0.374& 0.491 &0.597\\
$\quad$w/o SSQR& 0.401&	0.322	&0.441	&0.589\\
$\quad\quad\;\;\;\Delta$ ($\downarrow$)&{\cellcolor{red!20}10.69\%}&	{\cellcolor{red!20}13.90\%}	&{\cellcolor{red!20}10.18\%}	&1.34\%\\
\bottomrule
\end{tabular}
}
\label{tab_abl_llm}
\end{table}

\begin{figure}[t]
\centering


\includegraphics[width=0.9\linewidth]{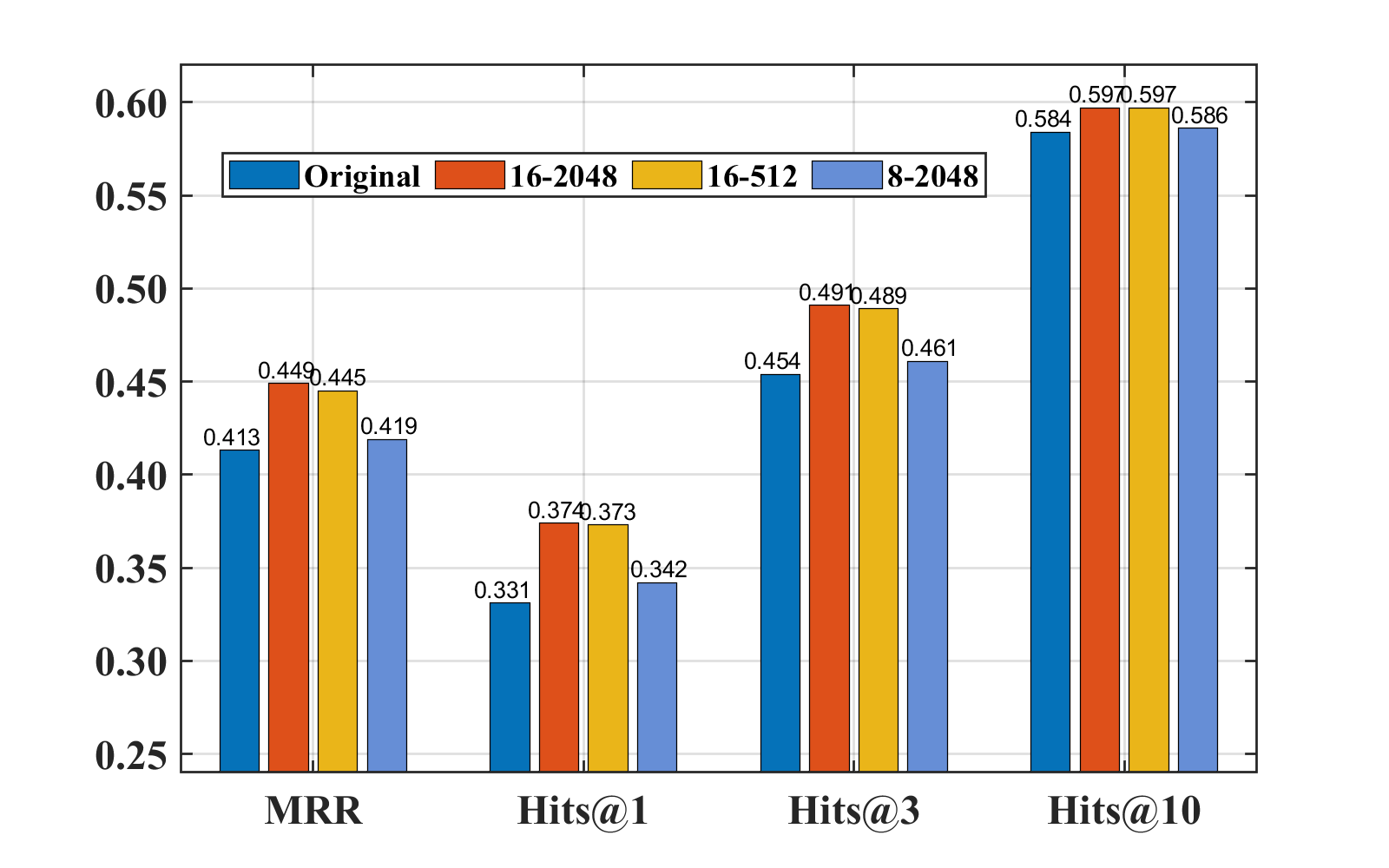}
\setlength{\abovecaptionskip}{0.1cm}
\setlength{\belowcaptionskip}{-0.3cm}
\caption{The impacts of quantized representation for KG link prediction task using LLMs on FB15k-237.
}
\label{fig_llm_mn}
\end{figure}

\noindent \textbf{Impacts of $M$ and $N$ for LLM Tuning.}
We explore the impacts of $M$ and $N$ for LLM tuning, the results are shown in Figure~\ref{fig_llm_mn}.
First, we present the results of \emph{Original}, which are the original results of AdaProp.
It is shown that all other results are better than those of AdaProp, showing it is effective for LLM fine-tuning with quantized representations.
The settings with $N$=16 and $M$=2048 have better results compared to 16-512 and 8-2048, indicating large values are needed to represent entity structural and semantic information, serving better features for LLMs.
$N$ is more important than $M$, which drops more performance, especially on the FB15k-237 dataset (16-512 even not dropping a lot), indicating the long quantized feature is beneficial for LLMs to distinguish entities.

\begin{figure}[t]
\centering
\includegraphics[width=0.9\linewidth]{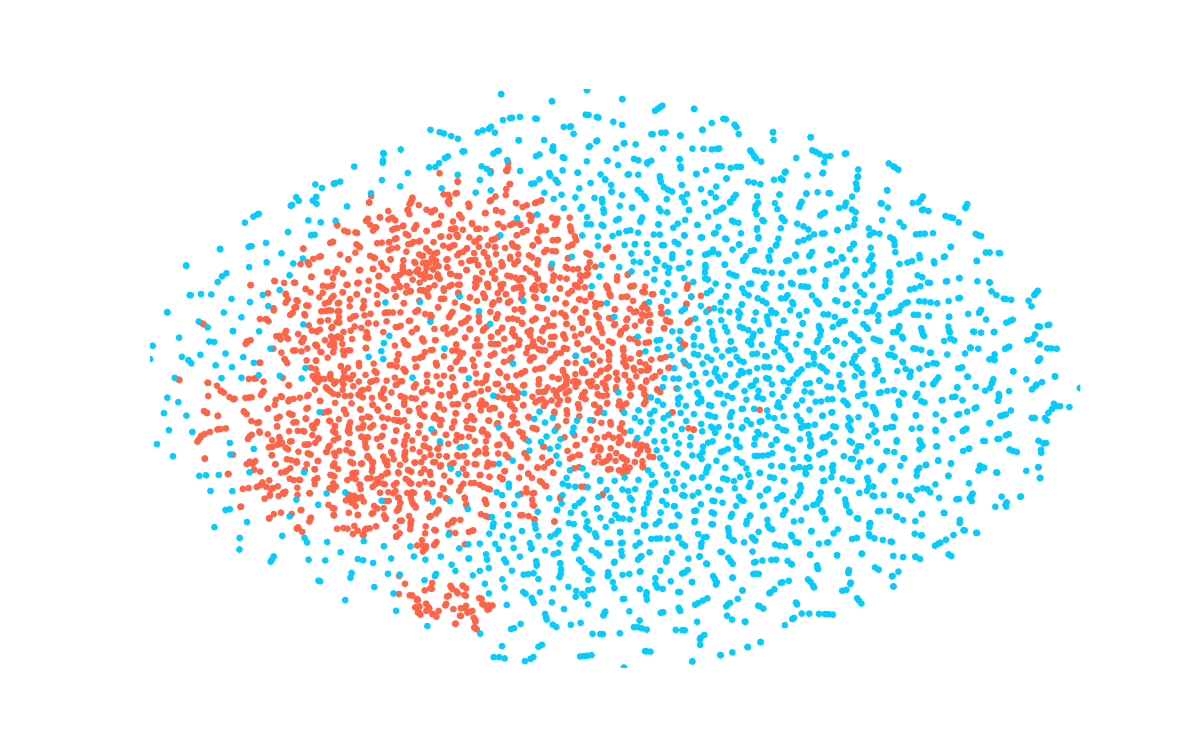}
\setlength{\abovecaptionskip}{-0.3cm}
\setlength{\belowcaptionskip}{-0.3cm}
\caption{Token embedding virtualization in LLMs (WN18RR dataset), where red and blue dots are real word tokens and code tokens, respectively.}
\label{fig_scatter1}
\end{figure}

\noindent \textbf{Token Embeddings in LLMs.}
To view the representation of codewords and actual language tokens in LLMs, we display all 2048 codewords and correspondingly sample an equal number of language tokens.
Further, we reduce the embeddings to 2-dimensional space using t-SNE~\cite{van2008visualizing} and plot the results in Figure~\ref{fig_scatter1}.
It is shown that these two types of tokens are generally divided into two categories, indicating they have different representation zones.
It is consistent with the intuition and indicates LLM can perceive that they are different types of tokens, showing potential for further exploration of LLM on KG tasks using SSQR.

\section{Related Work}
\label{sec_rel}

For parameter-efficient embeddings on large KGs, NodePiece~\cite{DBLP:conf/iclr/0001DWH22} introduces an anchor-based method to learn a fixed-size entity vocabulary, where unsupervised strategies of Personalized PageRank~\cite{page1999pagerank}, node degree, and random are used for anchor selection.
Each entity can be represented through \emph{k} closest anchors and their respective distances.
Further, EARL~\cite{DBLP:conf/aaai/ChenZYZGPC23} randomly samples 10\% entities as anchors and introduces connected relation information to match anchors' counterparts.
To simplify the whole process, \citet{DBLP:conf/emnlp/LiWLZM23} introduces random entity quantization (RandomEQ) to randomly set anchor entities and randomly select relations for matching.
The results show that RandomEQ achieves similar results compared to previous curated strategies and has more distinguishable ability.
In general, these methods are all in an unsupervised learning manner, which could be efficient for large KG embedding but fails to model comprehensive structural and semantic information.

Currently, numerous research studies are dedicated to incorporating KGs with LLMs to maximize and exploit their respective strengths~\cite{pan2024unifying}.
On one hand, using prompt engineering or retrieve strategies~\cite{DBLP:conf/emnlp/WeiH0K23,DBLP:conf/iclr/SunXTW0GNSG24,DBLP:journals/corr/abs-2407-06564}, the information of KGs be sampled and instantiated as tokens like natural language to input LLMs.
On the other hand, the triple-level or sub-graph structures of KG can be inputted to the LLMs to inject knowledge~\cite{hron2024training}.
However, because of the natural gap between the graph structure of KGs and the natural language,
how to seamlessly and effectively integrate the whole structural and semantic information of KGs with LLMs is an open problem.

\section{Conclusion and Potential Impacts}
\label{sec_con}
For seamlessly integrating KGs with LLMs, we introduce a self-supervised quantized representation method (SSQR).
It compresses the structural and semantic information of entities in KGs to a discrete permutation of codewords, which has a similar format as the natural language and can be directly inputted to the LLMs.
By specific instruction data and fine-tuning, LLMs can seamlessly learn KG's knowledge, which can be used in KG applications.
To verify the effectiveness of our method, we implement experiments on KG link prediction and triple classification tasks, which demonstrate the superiority of our method.
This innovative paradigm promises to usher in transformative techniques for KGs in the era of LLMs.
In the future, we will explore more applications and make progress towards unified frameworks for multiple KG tasks,
\eg, KG-based QA~\cite{DBLP:conf/acl/LuoETPG0MDSLZL24}, KG-based recommendation~\cite{DBLP:conf/kbse/HuangWXWCXL23}, and language modeling~\cite{DBLP:conf/naacl/LuoSZZ024}.


\section*{Limitations}

Despite our SSQR method's capacity to facilitate the seamless integration of KGs with LLMs,
our study encounters the generalization limitation due to the substantial computational burden associated with LLMs.
In most recent and our studies, LLMs are fine-tuned for a specific KG and the corresponding task,
which can not be applied to various KG tasks and largely limits the model generalization ability.
In the future, we will try to construct unified LLMs for KGs by implementing quantization within the same discrete space.


\bibliography{anthology}
\bibliographystyle{acl_natbib}

\hspace{5cm}
\appendix

\section{Statistics of WN18RR Dataset}

\begin{figure}[h]
\centering
\includegraphics[width=1.0\linewidth]{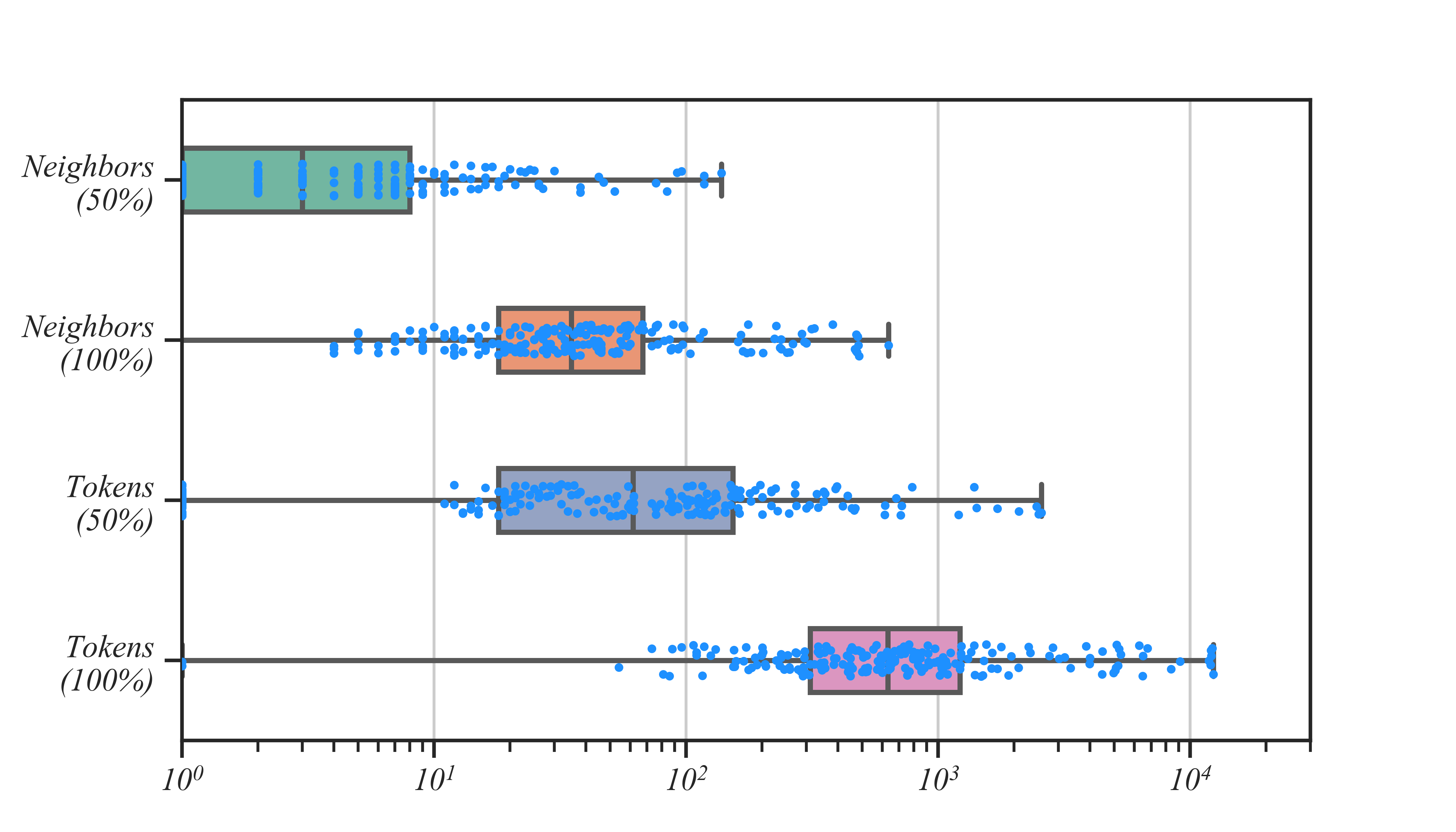}
\caption{The statistics of 2-hop sampled neighbors and needed tokens (by LLaMA2) for entities in WN18RR.}
\label{fig_intro_wn}
\end{figure}

Besides the statistic analysis of FB15k-237 in Figure~\ref{fig_intro_fb}, we also conduct the statistic analysis of WN18RR, which is shown in Figure~\ref{fig_intro_wn}.
Specifically, we sample 200 entities from the whole KG and there are two settings (50\% neighbor sampling and 100\% neighbors).
In the first setting of 50\%, the median and mean of neighbors are 4.0 and 10.37, while the median and mean number of needed tokens are 61.5 and 185.84, respectively.
For the setting of 100\%, the median and mean of neighbors are 33.5 and 79.05, while the median and mean number of needed tokens are 623.5 and 1492.74, respectively.
Compared to our SSQR, which only requires 16 tokens to represent each entity, both 50\% and 100\% settings demand a considerably higher number of tokens.

\section{Baselines}

In this section, we give detailed descriptions of various baselines utilized in the paper.

\subsection{Quantized Representations for KGs}


\quad$\bullet$\;NodePiece~\cite{DBLP:conf/iclr/0001DWH22}: The selection of quantized anchors relies on unsupervised strategies, including Personalized PageRank, node degree, and random approaches.

$\bullet$\;EARL~\cite{DBLP:conf/aaai/ChenZYZGPC23}: It randomly samples 10\% entities as quantized anchors and introduces connected relation information to match anchors’ counterparts.

$\bullet$\;Random entity quantization (RandomEQ for short)~\cite{DBLP:conf/emnlp/LiWLZM23}: It randomly sets anchor entities and randomly selects relations for matching. 

\subsection{KG Link Prediction}

\quad$\bullet$\;TransE~\cite{DBLP:conf/nips/BordesUGWY13}: The strategy of incorporating translational distance is utilized for learning representations of entities and relations.

$\bullet$\;CompGCN~\cite{DBLP:conf/iclr/VashishthSNT20}: Several entity-relation composition operations are proposed to combine the semantic information of neighbor entity-relation pairs in GNNs.

$\bullet$\;AdaProp~\cite{zhang2023adaprop}: An adaptive propagation path is learned to filter out irrelevant entities while preserving promising targets in the GNN framework.

$\bullet$\;MA-GNN~\cite{DBLP:conf/acl/XuBL23}: A dual-branch, multi-attention-based GNN model is employed to develop expressive entity representations.

$\bullet$\;TCRA~\cite{DBLP:conf/acl/Guo0LXN24}: A neuro-symbolic method that combines topological context learning with rule augmentation.

$\bullet$\;DiffusionE~\cite{DBLP:conf/kdd/CaoLWL24}: Introducing  diffusion process to KG embedding method.

$\bullet$\;KICGPT~\cite{DBLP:conf/emnlp/WeiH0K23}: 
The method utilizes a model based on embeddings as the retriever, which generates a ranked list of potential entities. An in-context learning strategy is then designed to guide ChatGPT in re-ranking these entities through multi-round interactions.

$\bullet$\;CSProm-KG-CD~\cite{DBLP:conf/eacl/LiTCL24}: 
It converts compact and structured triples into segments enriched with context by LLMs.
Following this, two custom auxiliary tasks (reconstruction and contextualization) are presented, which enable compact KGC models to incorporate insights derived from these enhanced triples.



$\bullet$\;ARR~\cite{chen2024new}:
A three-step (alignment, reasoning, and reranking) process designed to support and amplify conventional KG embedding models, without necessitating fine-tuning. The results are taken from the setting of LLAMA3-70B and RotatE~\cite{DBLP:conf/iclr/SunDNT19} model.

$\bullet$\;KG-FIT~\cite{DBLP:journals/corr/abs-2405-16412}: 
It involves the automatic construction of a semantically consistent entity hierarchy through clustering and LLM-guided refinement.
It also details a fine-tuning technique that incorporates knowledge from the hierarchical structure and pre-trained text embeddings of entities, thereby improving KG embeddings.
The results are from the HAKE model setting.

$\bullet$\;MKGL~\cite{DBLP:journals/corr/abs-2410-07526}:
A context retriever is introduced to help LLMs be aware of the textual and relational context of KGs.
A score retriever is also used to provide the score information.
LLaMA2 (7B) is utilized as the base LLM.


\subsection{KG Triple Classification}

\quad$\bullet$\;TransE~\cite{DBLP:conf/nips/BordesUGWY13}: The strategy of incorporating translational distance is utilized for learning representations of entities and relations.

$\bullet$\;DistMult~\cite{DBLP:journals/corr/YangYHGD14a}: It utilizes the semantic matching strategy, where the validity of a fact is depicted as the matching degree between the representation of entity and relation.

$\bullet$\;RotatE~\cite{DBLP:conf/iclr/SunDNT19}: It defines each relation as a rotation from the source entity to the target entity in a complex vector space.

$\bullet$\;Alpaca$_{\textsl{zero-shot}}$: It carries out zero-shot reasoning with Alpaca~\cite{taori2023stanford} with textual sequences for predicting the validity of a triple.

$\bullet$\;GPT-3.5$_{\textsl{zero-shot}}$: It carries out zero-shot reasoning with GPT-3.5~\footnote{https://openai.com/index/gpt-3-5-turbo-fine-tuning-and-api-updates/} with textual sequences for predicting the validity of a triple.

$\bullet$\;KG-LLaMA~\cite{DBLP:journals/corr/abs-2308-13916}: It carries out instruction tuning with LLaMA with textual sequences for predicting the validity of a triple.

$\bullet$\;KG-Alpaca~\cite{DBLP:journals/corr/abs-2308-13916}: It carries out instruction tuning with Alpaca with textual sequences for predicting the validity of a triple.

$\bullet$\;KoPA~\cite{DBLP:conf/mm/00090GX0C24}: It proposes a knowledge prefix adapter to effectively integrate pre-trained KG structural embeddings with LLMs.
Alpaca-7B is utilized as the LLM backbone.

\section{Experimental Details}
\label{app_details}

\begin{table}[h]
\setlength\tabcolsep{2pt} 
    \centering
    \caption{The statistics of WN18RR, FB15k-237, and FB15k-237N datasets. The former two are for link prediction. FB15k-237N dataset is for triple classification, where `/' splits the positive and negative samples.}
    \resizebox{0.48\textwidth}{!}{
    \begin{tabular}{lccccc}
        \toprule
        Dataset  & Ent & Rel &  Train&Valid & Test\\
        \midrule
        WN18RR& 40943 &11& 86835& 3034& 3134\\
        FB15k-237 &14541& 237& 272115& 17535& 20466\\
        \midrule
        FB15k-237N &13104& 93& 87282& 7041/7041& 8226/8226\\
        \bottomrule
    \end{tabular}
    }
    \label{tab_datasets}
\end{table}

The statistics of utilized datasets are shown in Table~\ref{tab_datasets}.
For the SSQR learning, the default embedding dimension is set to 200.
The GCN layer and dropout rate are 2 and 0.2.
The training batch is 1024.
For optimization, the learning rate is 0.0005 and the L2 regularization weight is 1e-8.
For LLM tuning, we utilize 4 NVIDIA H100 GPUs and the learning rate is set to 2e-5 with 3\% warmup ratio.
In the link prediction experiment, we first tune LLMs on the instruction data of CompGCN's training split to initialize.
Then, inspired by \citet{DBLP:conf/emnlp/WeiH0K23} and \citet{liu2024finetuning}, we divide the valid set into two segments in a 9:1 ratio.
The larger part is utilized to finetune LLMs to learn the ranking preference, while the smaller part is used for validation.
In the triple classification experiment, we only update the embedding layer and the last four Transformer layers of LLMs for tuning efficiency.
Meanwhile, $M$ and $N$ are set to 1024 and 16.
In the training instruction data, we randomly select negative samples at a rate 16 times of positive ones.
The instruction format of triple classification is shown in Table~\ref{tab_instruction_class}.

\begin{table}[h]
    \centering
    \begin{tabular}{c}
                \begin{tcolorbox}[colback=gray!10,
                      colframe=black,
                      width=7.cm,
                      boxrule=0.8pt,
                      arc=1mm, auto outer arc,
                      left = 1mm, 
                        right = 1mm,
                        top = 1mm,
                        bottom = 1mm,
                     ]
                \footnotesize{
                \textbf{Instruction:} Given a triple in the knowledge graph, you need to predict its validity based on the triple itself and entities' quantized representations.\\
                \textbf{Input:} The triple is: (\emph{h}, \emph{r}, \emph{t})\\
                The quantized representation of entity \emph{h} is: \blue{[Code(\emph{h})]}\\
                The quantized representation of entity \emph{t} is: \blue{[Code(\emph{t})]}\\
                Please determine the validity of the triple and respond True or False.\\
                \textbf{Output:} True/False
                }
                \end{tcolorbox} \\
    \end{tabular}
    \caption{Instruction format for triple classification.}
    \label{tab_instruction_class}
\end{table}

\section{Entropy and Jaccard Distance}
As presented by \citet{DBLP:conf/emnlp/LiWLZM23}, it is significant for the ability to distinguish different entities for quantized representations. We follow this study to calculate the entropy at the overall representation level and the Jaccard distance at the codeword-selection level.
The greater entropy and Jaccard distance values denote the greater distinguishable ability.
The entropy is calculated as:
\begin{equation}
    H=-\sum p\big(\textsl{Code}(e)\big)\cdot\log p\big(\textsl{Code}(e)\big).
\end{equation}
$p\big(\textsl{Code}(e)\big)$ is the relative frequency of quantized representation of entity $e$.
Moreover, the Jaccard distance is given by:
\begin{equation}
\small
    \mathcal{J}=\frac{1}{|\mathcal{E}|\cdot k}\sum_{e_i\in \mathcal{E}}\sum_{e_j\in \textsl{kNN}(e_i)} \!\!\!\!d\big(\textsl{Code}(e_i), \textsl{Code}(e_j)\big),
\end{equation}
\begin{equation}
\small
\begin{split}
    &d\big(\textsl{Code}(e_i), \textsl{Code}(e_j)\big)\\
    &=\frac{|\textsl{CSet}(e_i)\cup\textsl{CSet}(e_j)|-|\textsl{CSet}(e_i)\cap\textsl{CSet}(e_j)|}{|\textsl{CSet}(e_i)\cup\textsl{CSet}(e_j)|},
\end{split}
\end{equation}
where $\textsl{kNN}(e_i)$ retrieves $k$ entities.
Each possesses codes that exhibit the nearest Jaccard distance to the codes of $e_i$.
$\textsl{CSet}(e)$ is the set of $\textsl{Code}(e)$ by removing the order information of codewords of the entity representations.

\section{Additional Experimental Analysis}

\subsection{Training Process of SSQR}

\begin{figure}[h]
\centering
\includegraphics[width=1.0\linewidth]{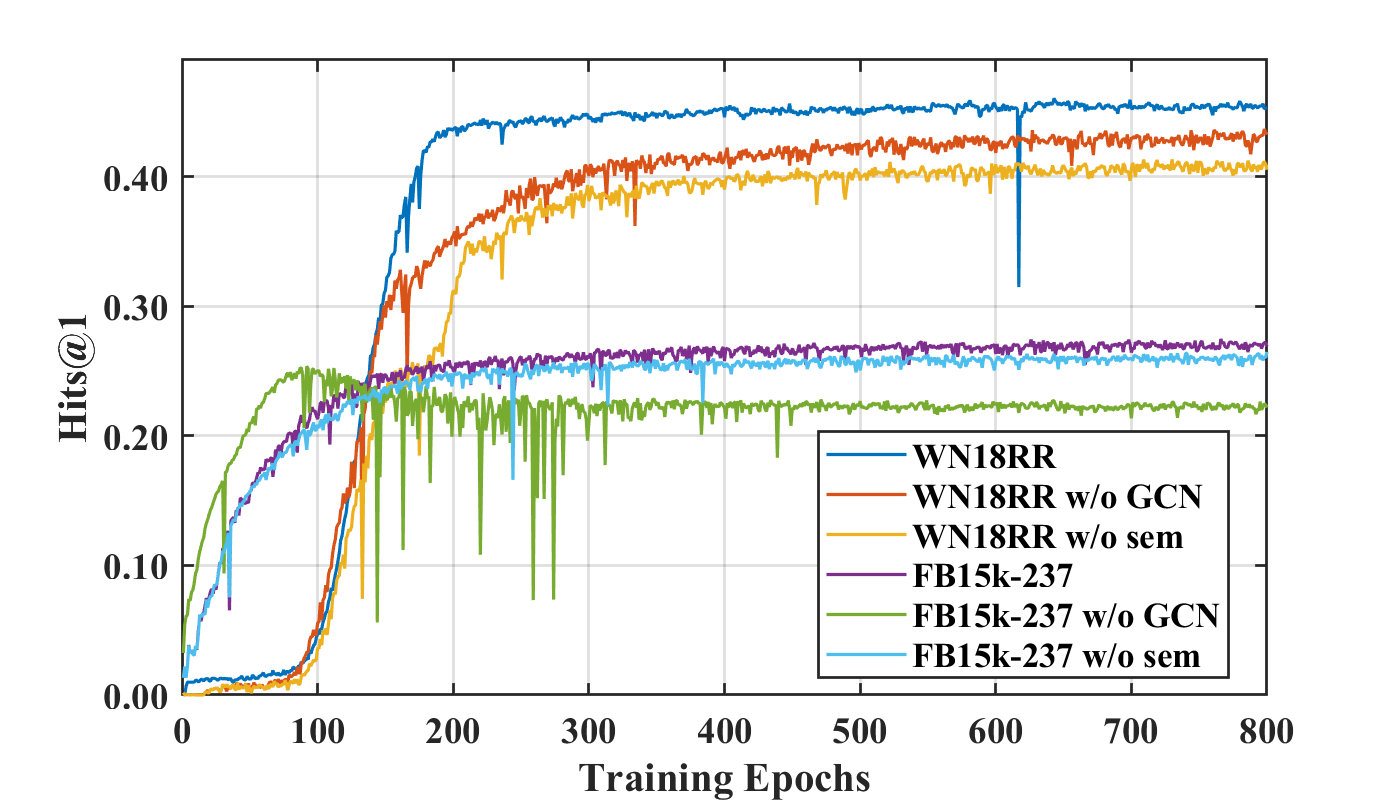}
\caption{The training process of SSQR, where the Hits@1 metric is used to show the model performance.}
\label{fig_training}
\end{figure}

We display the training process SSQR in Figure~\ref{fig_training}, where \emph{w/o GCN} and \emph{w/o sem} denote ablations for the structural embedding and semantic distilling, respectively.
The findings indicate that both structural embedding and semantic distilling contribute positively to the overall learning of quantized representation.
The influence of semantic information on the FB15k-237 dataset is less significant when compared to its effect on GCN.
Differently, semantic information is more important on the WN18RR dataset.
This could be attributed to the varying levels of KGs' sparsity.

\subsection{Relevance among Entity Codes on FB15k-237 Dataset}

\begin{figure*}[t]
\centering
\subfloat[Original text embedding.]{
\includegraphics[width=0.48\linewidth]{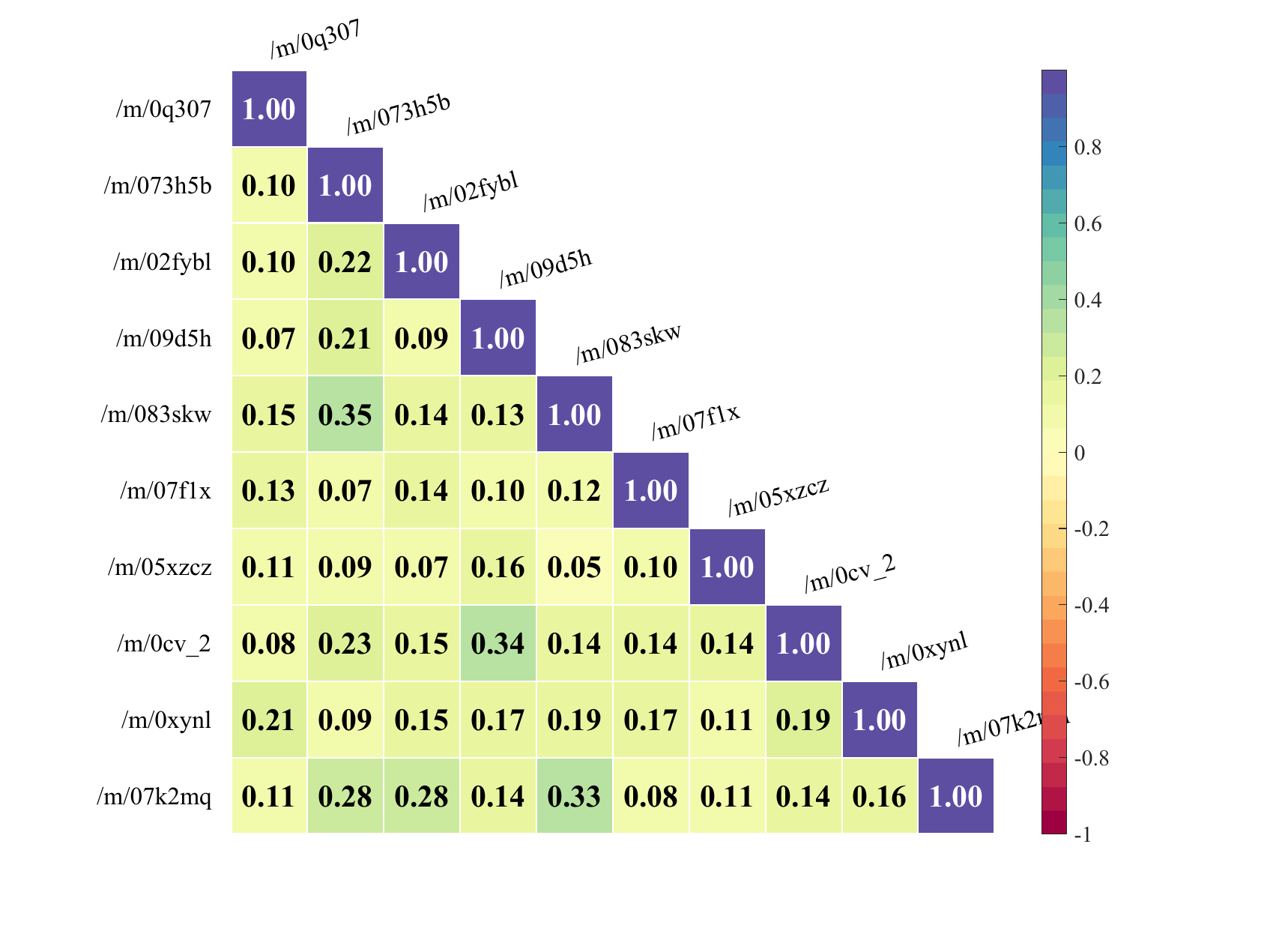}}
\hspace{-0.95cm}
\subfloat[SSQR.]{
\includegraphics[width=0.48\linewidth]{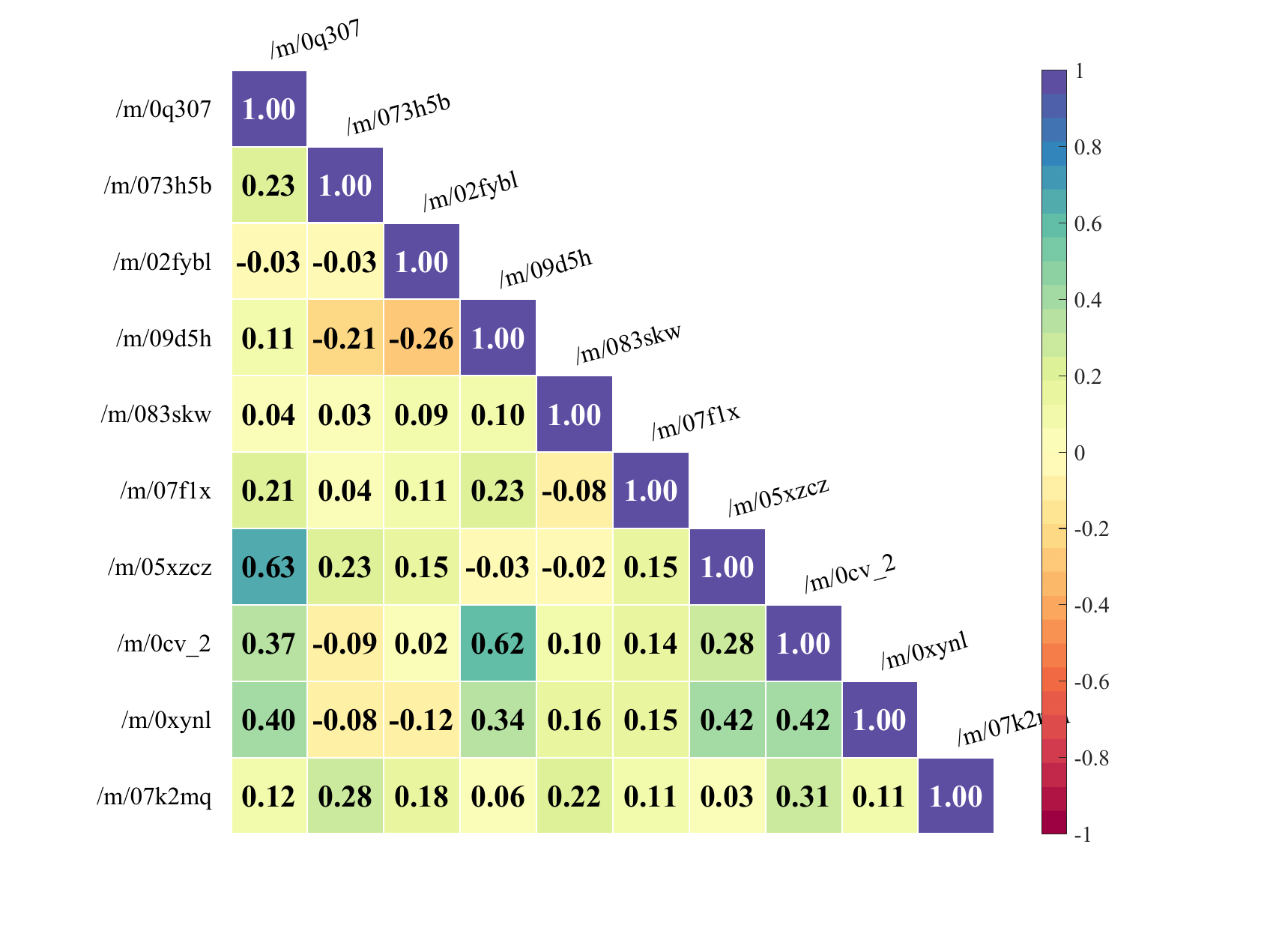}}

\subfloat[SSQR w/o GCN.]{
\includegraphics[width=0.48\linewidth]{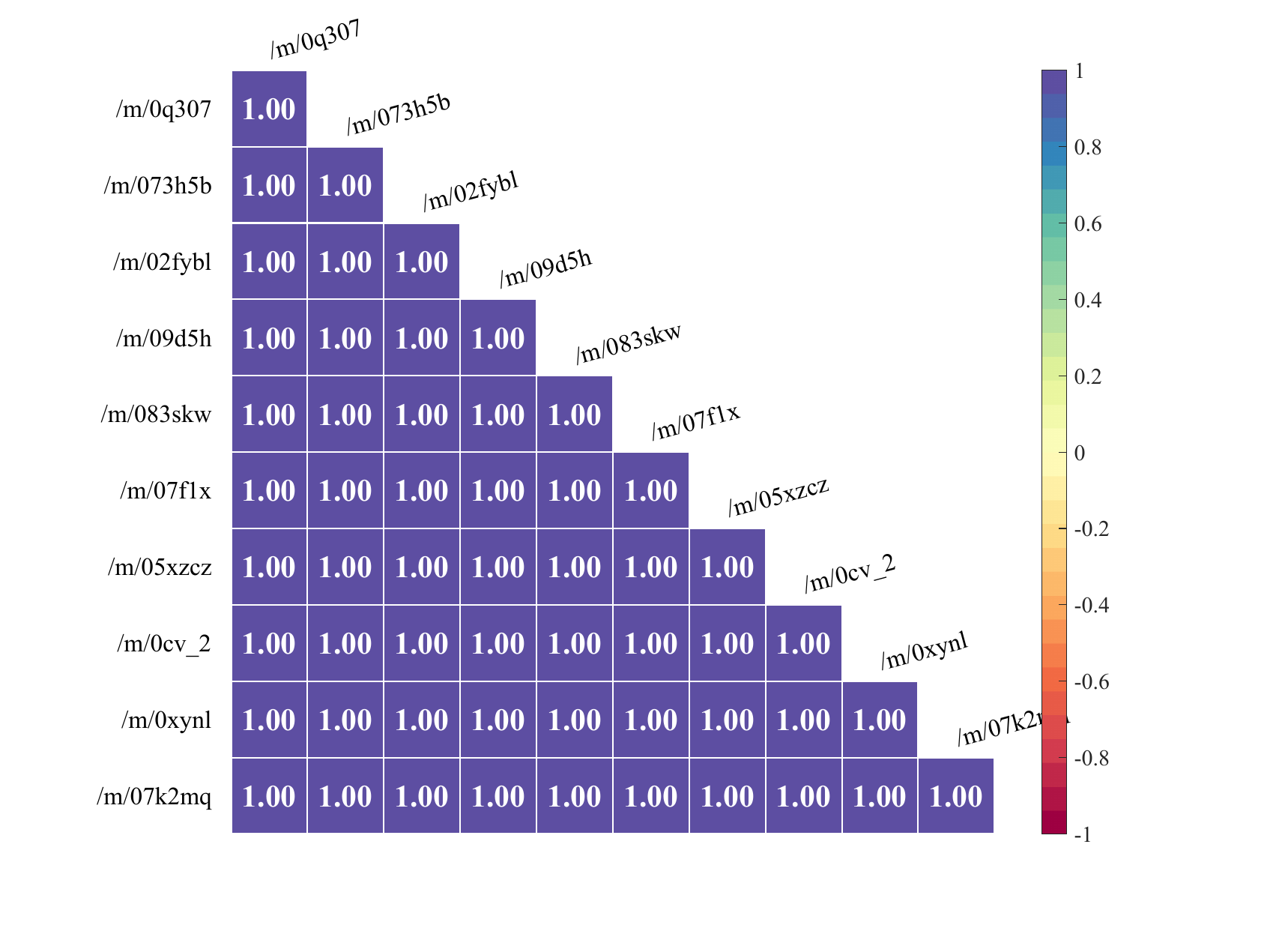}}
\hspace{-0.95cm}
\subfloat[SSQR w/o semantics.]{
\includegraphics[width=0.48\linewidth]{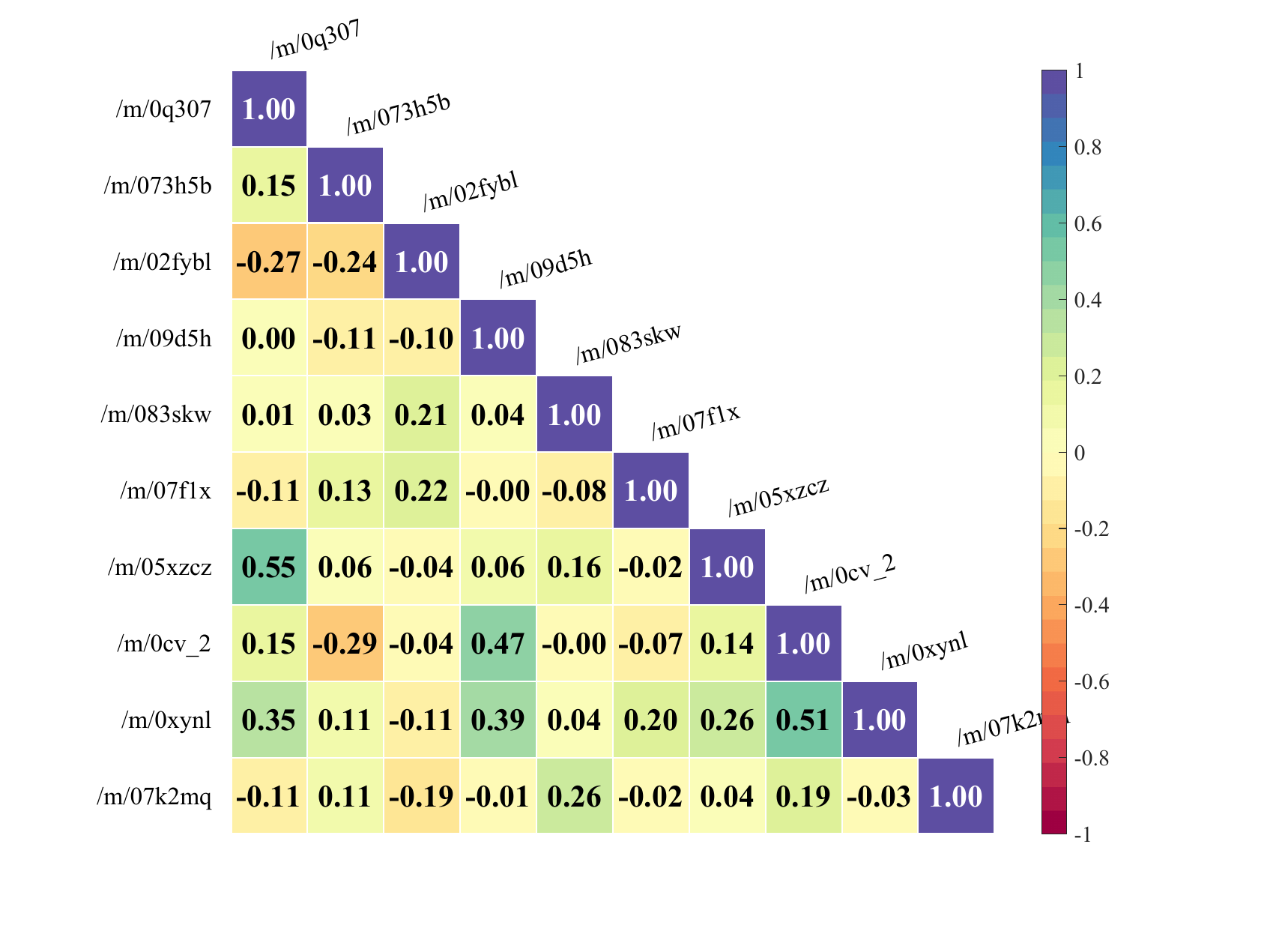}}
\caption{The cosine similarity of quantized representations on the FB15k-237 dataset (sampled 10 entities).}
\label{fig_simfb}
\end{figure*}

We also calculate the cosine similarity of quantized representation on the FB15k-237 dataset in Figure~\ref{fig_simfb}, which has the same setting as Figure~\ref{fig_simwn}.
The contents presented in these two figures are also similar.
When utilizing solely text embeddings, the corresponding similarities yield positive yet modest values. Moreover, the similarities associated with SSQR without the use of GCN are typically close to 1.
These observations suggest that entity representations occupy a limited portion of the existing space, thus failing to maximize the efficiency of representation. SSQR addresses this issue to some degree by providing a broader range and diversity of similarities.

\subsection{Impacts of $M$ and $N$ for LLM Tuning on FB15k-237 Dataset}

\begin{figure}[h!]
\centering


\includegraphics[width=0.9\linewidth]{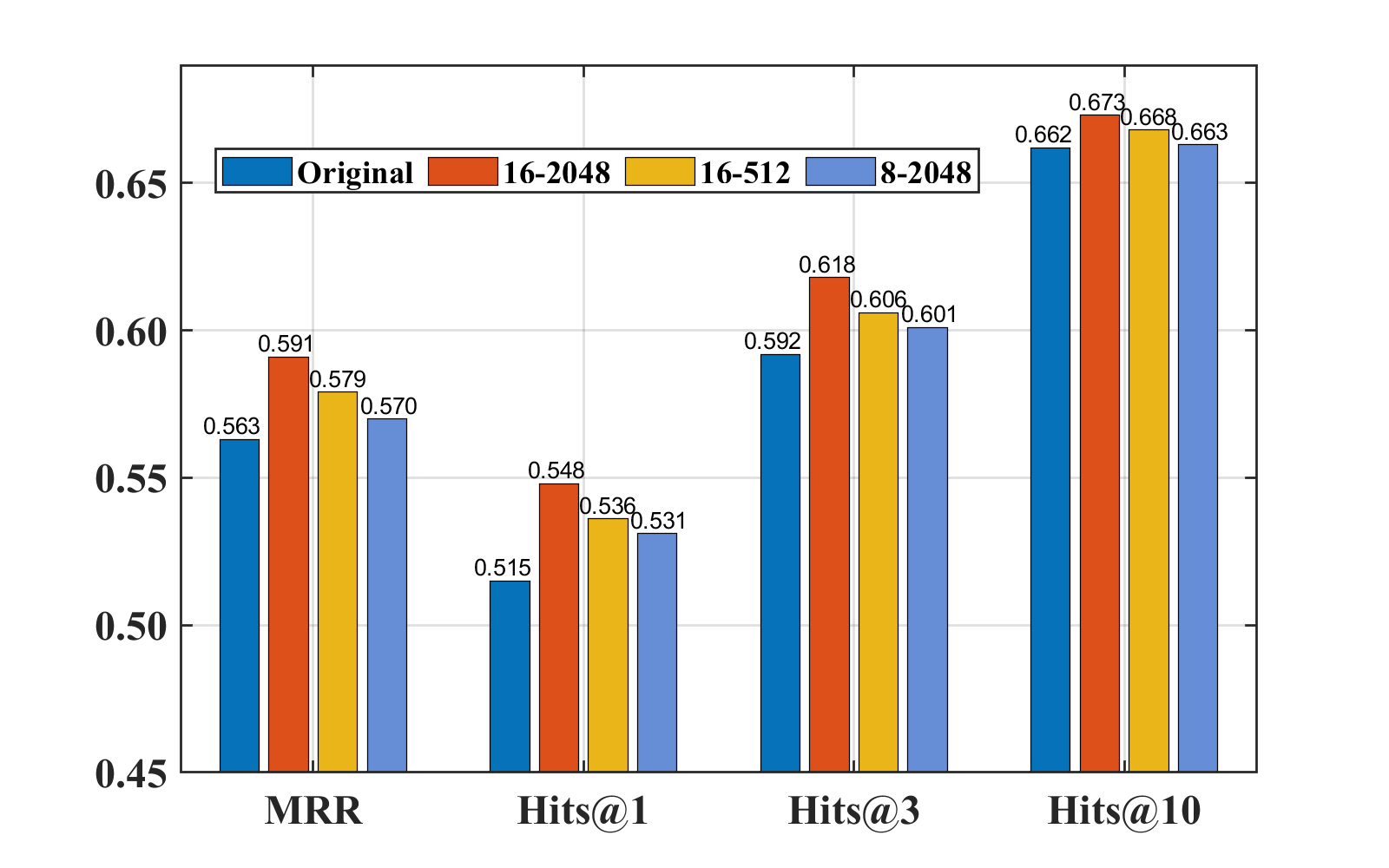}

\caption{The impacts of quantized representation for KG link prediction task using LLMs on WN18RR.
}
\label{fig_llm_mn_fb}
\end{figure}

We also explore the impacts of $M$ and $N$ for LLM tuning on the FB15k-237 dataset, the results are shown in Figure~\ref{fig_llm_mn_fb}.
It can lead to conclusions similar to Figure~\ref{fig_llm_mn}.

\subsection{Token Embeddings in LLMs on FB15k-237 Dataset}

\begin{figure}[h!]
\centering
\includegraphics[width=1.0\linewidth]{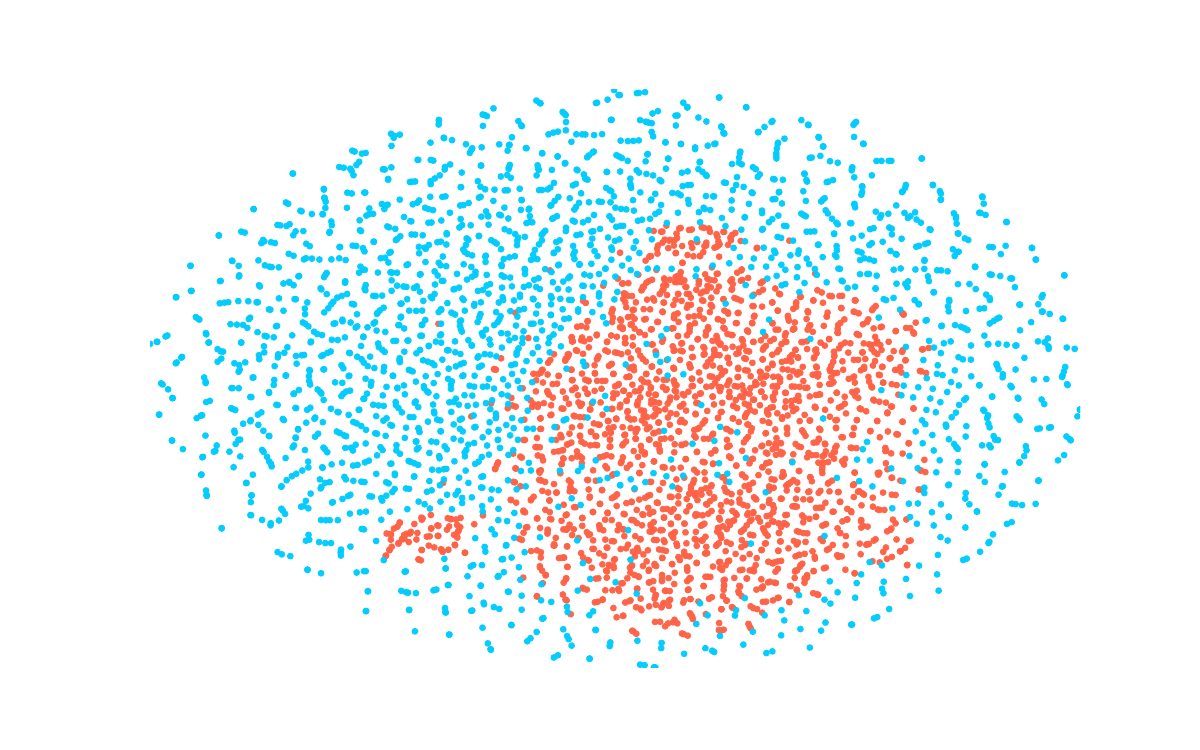}
\caption{Token embedding virtualization in LLMs (FB15k-237 dataset), where red and blue dots are real word tokens and code tokens, respectively.}
\label{fig_scatter2}
\end{figure}

Similar to Figure~\ref{fig_scatter1}, we display the real word tokens and learned code tokens using t-SNE in Figure~\ref{fig_scatter2}.
The evidence also suggests that these two types of tokens typically fall into distinct categories, implying they each have unique representation areas.

\subsection{Case Studies}

To intuitively show the seamlessly integrating KG tasks with LLMs, we carry out case studies in Table~\ref{tab_case1}, \ref{tab_case2}, and \ref{tab_case3}, covering both link prediction and triple classification tasks.
It demonstrates that our method can effectively address both tasks, indicating the validity and good generalization ability of our proposed SSQR.

\begin{table*}[]
    \centering
    \begin{tabular}{c}
                \begin{tcolorbox}[colback=gray!1,
                      colframe=black,
                      width=16cm,
                      boxrule=0.8pt,
                      arc=1mm, auto outer arc,
                      left = 1mm, 
                        right = 1mm,
                        top = 1mm,
                        bottom = 1mm,
                     ]
                \footnotesize{
                \textbf{Input:} This is a knowledge graph completion task, which needs to predict the tail entity for an incomplete query triplet.\\
                The query triplet is (radiotherapy, hypernym, ?).\\
                The quantized representation of entity radiotherapy is: [2006] [588] [350] [1486] [214] [929] [328] [1424] [1792] [919] [944] [740] [438] [843] [147] [628]\\
                The answer candidates and corresponding quantized representations are as follows:\\
                disease, [156] [1880] [1777] [185] [121] [720] [783] [1713] [945] [1077] [180] [1576] [1574] [1433] [216] [1280]\\
                tomography, [182] [597] [657] [1486] [404] [468] [732] [564] [833] [1470] [1756] [626] [1674] [843] [1928] [513]\\
                medical care, [422] [68] [1329] [1517] [1251] [431] [1479] [1445] [1666] [407] [952] [406] [1337] [388] [1982] [685]\\
                status, [1721] [1906] [1773] [1811] [12] [892] [1625] [1476] [1561] [176] [534] [1463] [1657] [368] [70] [1618]\\
                physiological state, [1721] [718] [267] [394] [120] [1105] [885] [1823] [1496] [23] [952] [406] [1559] [1198] [1149] [1800]\\
                medical science, [565] [413] [842] [1517] [350] [873] [575] [595] [721] [935] [1554] [175] [708] [1643] [1820] [1775]\\
                infection, [565] [1594] [990] [1066] [974] [40] [434] [874] [1401] [371] [1700] [1118] [1709] [52] [71] [1408]\\
                picturing, [788] [168] [641] [1797] [927] [711] [1608] [123] [1163] [1460] [952] [406] [1752] [1464] [553] [1158]\\
                medicine, [1879] [1216] [691] [296] [1743] [892] [1851] [595] [2039] [1428] [426] [740] [399] [579] [433] [1987]\\
                unhealthiness, [1389] [644] [570] [258] [635] [647] [732] [1139] [1660] [407] [464] [1020] [1574] [1905] [926] [1971]\\
                grounds, [1268] [1053] [803] [780] [1194] [285] [328] [289] [1163] [915] [1921] [1020] [524] [1774] [430] [1572]\\
                defense reaction, [1881] [1821] [1620] [1703] [435] [995] [908] [1308] [1596] [1598] [401] [2008] [903] [817] [92] [1158]\\
                radiology, [1478] [588] [1340] [1797] [1436] [1914] [1894] [1424] [634] [1460] [1756] [740] [673] [843] [108] [1088]\\
                radioscopy, [1005] [1002] [1441] [137] [1436] [1378] [1479] [1649] [1544] [1470] [534] [626] [902] [272] [904] [1874]\\
                treat, [396] [2007] [1935] [1305] [1993] [1030] [1690] [1445] [1203] [1417] [1554] [495] [1752] [1001] [1236] [98]\\
                specialize, [1005] [1933] [1976] [780] [927] [1728] [575] [105] [1791] [1598] [616] [1118] [1752] [425] [437] [1847]\\
                therapy, \blue{[396] [816] [81] [488] [336] [1164] [1690] [1288] [900] [915] [1554] [175] [666] [1622] [765] [685]}\\
                specialism, [384] [816] [599] [394] [435] [789] [1479] [105] [664] [407] [1554] [103] [1752] [1708] [697] [1130]\\
                symptom, [1721] [1913] [772] [858] [120] [1150] [1374] [289] [1666] [1417] [944] [2008] [1454] [958] [1169] [1800]\\
                medicine, [156] [350] [1599] [1955] [1368] [508] [1527] [1445] [1561] [1460] [426] [1142] [940] [653] [793] [471]\\
                Please generate quantized representations of the top-3 potential answer entities, ranked from highest to lowest: \\
                \textbf{LLM Output:} 1, \blue{[396] [816] [81] [488] [336] [1164] [1690] [1288] [900] [915] [1554] [175] [666] [1622] [765] [685]}\\2, [156] [1880] [1777] [185] [121] [720] [783] [1713] [945] [1077] [180] [1576] [1574] [1433] [216] [1280]\\3, [182] [597] [657] [1486] [404] [468] [732] [564] [833] [1470] [1756] [626] [1674] [843] [1928] [513]\\
                \textbf{Ground Truth:} \blue{[396] [816] [81] [488] [336] [1164] [1690] [1288] [900] [915] [1554] [175] [666] [1622] [765] [685]}
                }
                \end{tcolorbox} \\
    \end{tabular}
    \caption{Case study on WN18RR for link prediction using LLaMA2. The code of ground truth \emph{therapy} is ranked to the first position from 17-th.}
    \label{tab_case1}
\end{table*}

\begin{table*}[]
    \centering
    \begin{tabular}{c}
                \begin{tcolorbox}[colback=gray!1,
                      colframe=black,
                      width=16cm,
                      boxrule=0.8pt,
                      arc=1mm, auto outer arc,
                      left = 1mm, 
                        right = 1mm,
                        top = 1mm,
                        bottom = 1mm,
                     ]
                \footnotesize{
                \textbf{Input:} This is a knowledge graph completion task, which needs to predict the tail entity for an incomplete query triplet.\\
                The query triplet is (Valparaiso University, inverse relation of /location/location/contains, ?).\\
                The quantized representation of entity Valparaiso University is [527] [1345] [1849] [1227] [1751] [2038] [818] [515] [1417] [333] [29] [721] [1691] [798] [1033] [153]\\
                The answer candidates and corresponding quantized representations are as follows:\\
                Minnesota, [1532] [258] [1837] [357] [923] [1994] [638] [555] [771] [1003] [1736] [1473] [1495] [1436] [1313] [20]\\
                New York, [661] [1243] [542] [1741] [1907] [1799] [858] [1794] [1916] [458] [1844] [909] [438] [1737] [686] [963]\\
                California, [1059] [1286] [1604] [846] [1086] [451] [1087] [1794] [994] [297] [1463] [159] [556] [1836] [407] [963]\\
                Massachusetts, [202] [1243] [977] [757] [304] [389] [1172] [1308] [1916] [1858] [1323] [11] [841] [1680] [1798] [1885]\\
                Illinois, [961] [1025] [1267] [174] [643] [1951] [1742] [1794] [1720] [1481] [543] [1883] [695] [1921] [182] [963]\\
                New York City, [1458] [326] [1707] [239] [151] [640] [1366] [1794] [610] [458] [1844] [932] [122] [311] [121] [868]\\
                United Kingdom, [51] [193] [1354] [669] [1867] [881] [480] [1271] [392] [1858] [650] [909] [1503] [1126] [1550] [153]\\
                Pennsylvania, [361] [825] [1052] [1655] [1670] [732] [951] [1569] [275] [1995] [543] [4] [753] [351] [331] [637]\\
                Los Angeles, [1584] [1231] [1707] [1461] [1867] [1466] [265] [1933] [850] [1533] [805] [1128] [1824] [1823] [307] [963]\\
                Florida, [2016] [326] [542] [1614] [462] [1433] [1388] [819] [926] [1289] [1321] [563] [1977] [1144] [1268] [662]\\
                Ohio, [1643] [1889] [1604] [88] [1364] [485] [1819] [1569] [54] [1582] [1500] [411] [438] [125] [1636] [20]\\
                Texas, [2012] [1845] [1207] [412] [531] [1394] [1004] [688] [653] [1671] [1790] [1690] [1732] [1686] [1721] [1205]\\
                Virginia, [99] [825] [738] [1859] [1287] [1540] [708] [780] [653] [662] [756] [1873] [1514] [1686] [59] [409]\\
                England, [848] [1220] [1052] [590] [175] [451] [529] [1933] [808] [1598] [1790] [1988] [670] [524] [121] [136]\\
                Connecticut, [633] [1025] [672] [338] [1694] [1799] [1528] [1177] [1949] [458] [1703] [411] [395] [33] [233] [1013]\\
                Indiana, \blue{[497] [1875] [1849] [377] [1694] [61] [1471] [1445] [392] [1672] [1500] [300] [711] [1839] [331] [136]}\\
                Los Angeles County, [662] [1772] [558] [1623] [304] [1755] [1388] [1794] [1554] [333] [1662] [1258] [1474] [1764] [1021] [409]\\
                Louisiana, [1584] [326] [1309] [50] [1035] [0] [546] [728] [1130] [297] [1284] [11] [1514] [1929] [1248] [136]\\
                Michigan, [172] [326] [382] [174] [1364] [1774] [1165] [664] [994] [1480] [1923] [431] [816] [1587] [1443] [860]\\
                Hennepin County, [584] [1890] [1263] [1168] [1990] [656] [784] [1735] [102] [1533] [1979] [1005] [1312] [786] [873] [1294]\\
                Please generate quantized representations of the top-3 potential answer entities, ranked from highest to lowest: \\
                \textbf{LLM Output:} 1, \blue{[497] [1875] [1849] [377] [1694] [61] [1471] [1445] [392] [1672] [1500] [300] [711] [1839] [331] [136]}\\
                2, [1532] [258] [1837] [357] [923] [1994] [638] [555] [771] [1003] [1736] [1473] [1495] [1436] [1313] [20]\\
                3, [661] [1243] [542] [1741] [1907] [1799] [858] [1794] [1916] [458] [1844] [909] [438] [1737] [686] [963]\\
                \textbf{Ground Truth:} \blue{[497] [1875] [1849] [377] [1694] [61] [1471] [1445] [392] [1672] [1500] [300] [711] [1839] [331] [136]}
                }
                \end{tcolorbox} \\
    \end{tabular}
    \caption{Case study on FB15k-237 for link prediction using LLaMA2. The code of ground truth \emph{Indiana} is ranked to the first position from 16-th.}
    \label{tab_case2}
\end{table*}

\begin{table*}[]
    \centering
    \begin{tabular}{c}
                \begin{tcolorbox}[colback=gray!1,
                      colframe=black,
                      width=16cm,
                      boxrule=0.8pt,
                      arc=1mm, auto outer arc,
                      left = 1mm, 
                        right = 1mm,
                        top = 1mm,
                        bottom = 1mm,
                     ]
                \footnotesize{
                \textbf{Input:} Given a triple in the knowledge graph, you need to predict its validity based on the triple itself and entities' quantized representations.\\
                The triple is: (Carl Foreman, /people/person/place\_of\_birth, Chicago)\\
                The quantized representation of entity ``Carl Foreman'' is: \blue{[72] [114] [150] [336] [519] [494] [155] [482] [594] [934] [1012] [677] [473] [450] [827] [455]}\\
                The quantized representation of entity ``Chicago'' is: \blue{[603] [413] [385] [937] [731] [735] [836] [92] [400] [372] [563] [225] [82] [526] [81] [229]}\\
                Please determine the validity of the triple and respond True or False.\\
                \textbf{LLM Output:} True\\
                \textbf{Ground Truth:} True\\
                \\
                \textbf{Input:} Given a triple in the knowledge graph, you need to predict its validity based on the triple itself and entities' quantized representations.\\
                The triple is: (Jessica Lange, /people/person/profession, Lawyer)\\
                The quantized representation of entity ``Jessica Lange'' is: \blue{[23] [712] [484] [202] [289] [211] [772] [667] [870] [903] [213] [693] [360] [59] [868] [722]}\\
                The quantized representation of entity ``Lawyer'' is: \blue{[760] [84] [976] [802] [430] [735] [912] [480] [966] [411] [284] [113] [727] [744] [333] [56]}\\
                Please determine the validity of the triple and respond True or False.\\
                \textbf{LLM Output:} False\\
                \textbf{Ground Truth:} False
                }
                \end{tcolorbox} \\
    \end{tabular}
    \caption{Two cases on FB15k-237N dataset for triple classification using LLaMA2.}
    \label{tab_case3}
\end{table*}

\end{document}